\documentclass{article}

\usepackage[utf8]{inputenc} 
\usepackage[T1]{fontenc}    
\usepackage{url}            
\usepackage{booktabs}       
\usepackage{amsfonts}       
\usepackage{nicefrac}       
\usepackage{microtype}      
\usepackage{xcolor}         
\usepackage{graphicx}
\usepackage{amsmath}
\usepackage{amssymb}
\usepackage{algorithm}
\usepackage{algorithmic}
\usepackage{tcolorbox}
\usepackage{listings}
\usepackage{hyperref}       
\usepackage{titlesec}
\usepackage{xspace}
\usepackage{mathtools}
\usepackage{bbm}
\usepackage{enumitem}
\usepackage{caption}
\usepackage{subcaption}
\usepackage{multirow}
\usepackage{pdflscape}
\usepackage[toc,page,header]{appendix}
\usepackage{minitoc}
\usepackage{amsthm}
\usepackage[capitalise, nameinlink]{cleveref}
\usepackage{tabularx}
\usepackage{wrapfig}
\usepackage{etoc}

\newcolumntype{Y}{>{\centering\arraybackslash}X}

\newcommand{\ACRO}{\textit{Visible Touch}\xspace}

\usepackage[preprint]{corl_2026} 

\usepackage{siunitx}

\definecolor{mydarkblue}{rgb}{0,0.08,0.45}

\hypersetup{
    pdftitle={Visible Touch: Rendering Contact for Visuomotor Policies},
    pdfauthor={Metin Alp Dogan, Edward Sun, Feng Xu, Daniel Wu, Allen Peng, Dennis Hong, Yuchen Cui},
    colorlinks=true,
    linkcolor=mydarkblue,
    filecolor=mydarkblue,
    urlcolor=mydarkblue,
    citecolor=mydarkblue,
}

\title{
\Large \ACRO: Rendering Contact for Visuomotor Policies} 

\makeatletter
\def\thanks#1{\protected@xdef\@thanks{\@thanks
        \protect\footnotetext{#1}}}
\makeatother

\author{
  Metin Alp Dogan$^{*}$,
  Edward Sun$^{*}$,
  Feng Xu$^{*}$,
  Daniel Wu,
  Allen Peng, \\[5pt]
  \textbf{Dennis Hong,
  Yuchen Cui}
  \thanks{{
\noindent  $^{*}$ Equal contribution. 
}}
  \\[5pt]
  University of California, Los Angeles 
}

\begin{document}
\maketitle

\doparttoc 
\faketableofcontents 
\part{} 
\vspace{-1.6cm}

\begin{abstract}
Integrating contact information into visuomotor policies remains an
open problem. Touch is essential to robust manipulation, yet most
modern policies, including pretrained vision-language-action (VLA)
models, operate from vision and proprioception alone. Existing
approaches to closing this gap require specialized tactile hardware,
add separate tactile encoders, or commit to non-image policy
backbones, all incompatible with the modern paradigm of
image-conditioned policies built on pretrained 2D visual
representations. Our key insight is that the bottleneck is not the
contact information itself, but how it is delivered: when contact
signals are exposed in the same spatial frame as the scene the
policy already attends to, they become directly usable by any
image-conditioned policy without architectural changes. We
operationalize this insight in \ACRO, paired with a custom low-cost
magnetic contact sensor that is open-sourced and fabricated from
off-the-shelf parts via a parametric CAD-to-mold pipeline.
Across the LIBERO benchmark, \ACRO\ improves BC-Transformer success by 15.7 percentage points on average in the 2-view setting, with similar gains in the 1-view setting; controlled comparisons show that the contact-integration strategy strongly affects how effectively tactile information is used. The pattern holds when fine-tuning pretrained VLAs: miniVLA on LIBERO gains 25 percentage points on
average, and $\pi_{0.5}$ on four real-world contact-rich tasks
gains 30 percentage points with our custom sensor. Project website: \url{https://visibletouch.github.io/}
\end{abstract}



\section{Introduction}
\label{sec:intro}

Touch is fundamental to human manipulation. We rely on it to know
whether we have grasped an object securely, when contact has been
made or broken, and how firmly we are gripping. Visuomotor robot
policies, by contrast, overwhelmingly operate from vision and
proprioception alone, even on tasks where contact information would
be most diagnostic. A growing body of work has explored how to
integrate contact sensing into learned policies, but no approach has
emerged as a clear default; modern vision-language-action (VLA)
models continue to leave contact out of their input pipelines
despite increasing attention to contact-rich manipulation.

This difficulty stems from three challenges. The first is
\textit{sensing}: high-fidelity tactile sensors are typically
expensive, bulky, or require specialized fabrication, putting them
out of reach for many setups. The second is \textit{signal
interpretation}: raw tactile readings are difficult to interpret in
isolation, since a contact measurement means different things
depending on the gripper's pose, the task phase, and the object in
hand. The third is \textit{architectural integration}, the
bottleneck for modern policies: existing approaches either
concatenate contact to the proprioceptive state
vector~\cite{xie2024justaddforce, liu2025factr} or add a separate
tactile encoder that processes contact as a parallel input
stream~\cite{xu2025manifeel, saka2026contact, huang2025tactilevla},
both modifying the policy architecture and often requiring
retraining or extending the visual encoder, which makes them
difficult to scale to pretrained visuomotor policies.

\begin{figure}[ht]
    \centering
    \includegraphics[width=\linewidth]{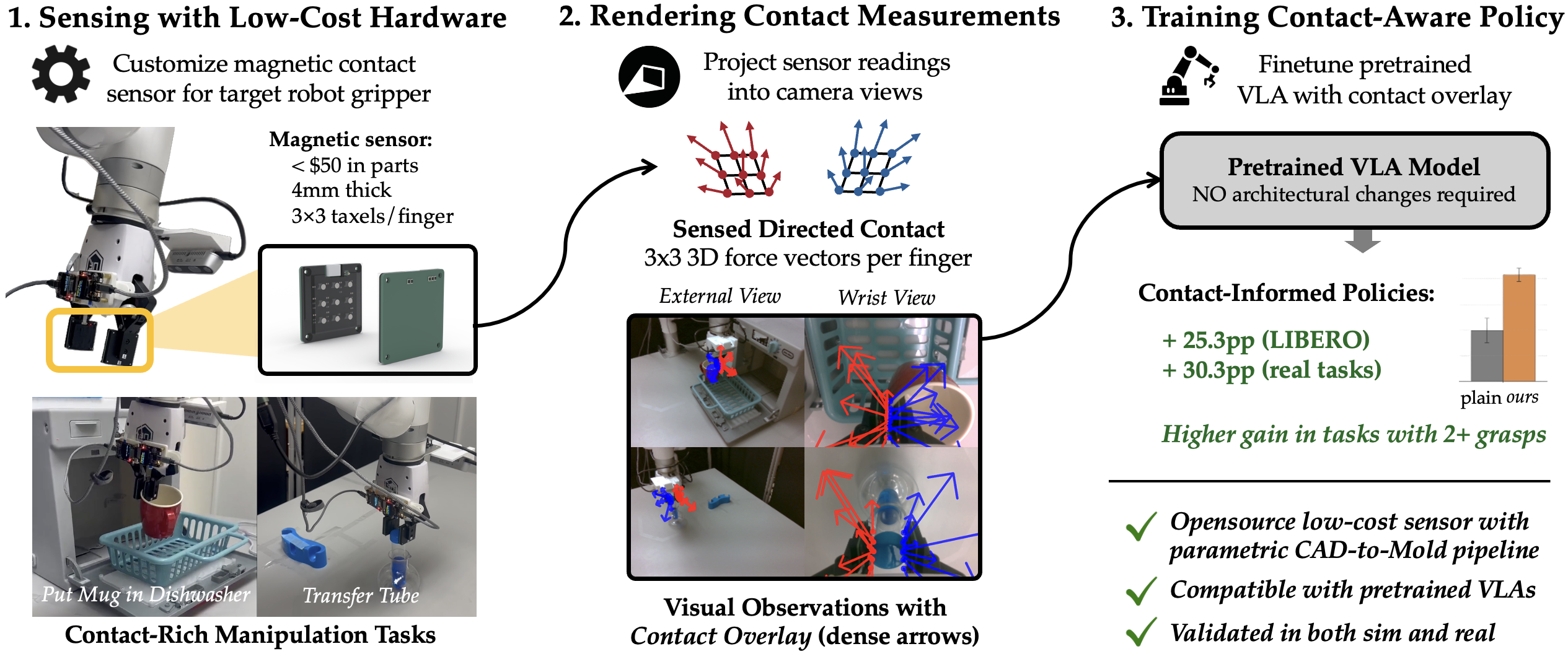}
    \caption{\small \textbf{Overview of \ACRO.}
    A low-cost magnetic contact sensor captures per-taxel contact,
    which we render as an image-space overlay on the policy's RGB
    observations. The augmented images are drop-in inputs for
    fine-tuning pretrained VLAs, requiring no architectural changes.
    Contact overlays improve average task success by $+25.3 pp$ on
    LIBERO fine-tuning suites and $+30.3 pp$ on real-world
    contact-rich manipulation tasks.}
    \label{fig:overview}
    \vspace{-1em}
\end{figure}
\normalsize

These challenges are usually addressed independently: cheaper
hardware for sensing, learned calibration or fusion models for
signal interpretation, custom architectures for integration. We
observe a common resolution: an image-conditioned policy already
attends to spatial structure in its RGB input, so if contact is
rendered into that same image, it inherits the policy's spatial
reasoning for free — no separate processing pathway, no calibrated
forces. This motivates a simpler alternative: \textbf{render
contact information directly into the policy's image input as a
spatial overlay}.

We introduce \ACRO, a method and hardware package for image-space
contact-aware policy learning (\cref{fig:overview}). The overlay is
computed from raw sensor readings and camera extrinsics, then
composited onto the existing RGB observations. Because contact
arrives as pixels in the image the policy already consumes, the
method serves as a drop-in augmentation for any image-conditioned
policy, including pretrained VLAs. Moreover, normalization and
visual scaling absorb absolute sensor scale, so the sensor need
only respond monotonically to contact rather than report calibrated
forces — permitting a low-cost open-source magnetic sensor paired
with a parametric CAD-to-mold pipeline that generates molds from
user-specified geometry. 

Conceptually, \ACRO is the 2D-image
counterpart of work that geometrically grounds tactile signals as
3D points~\cite{yuan2023robotsynesthesia, huang20243dvitac}; while
those methods commit to point-cloud backbones, ours targets the 2D
image-token architectures dominating modern manipulation policies.
The same principle of rendering auxiliary state into the visual
input has been explored for end-effector
pose~\cite{dai2025aimbot}, trajectory
sketches~\cite{gu2024rttrajectory}, and motion
history~\cite{zheng2024tracevla}, but never for contact: a signal
that is sparse, event-driven, and visually hidden at the moments it
matters most.

We evaluate \ACRO\ on both simulated and real tasks. Across all
four LIBERO suites, image-space overlays improve BC-Transformer success rates by 15.7 percentage points in the 2-view setting and 12.5 percentage points in the 1-view setting, with the largest gains ($+43$ $pp$) on long-horizon multi-grasp tasks. The pattern holds when fine-tuning pretrained VLAs: miniVLA on LIBERO gains 25 percentage points on average, and $\pi_{0.5}$ on four real-world contact-rich tasks gains 30 percentage points with our custom magnetic contact sensor, showing that the benefit transfers across architectures and
from simulation to real-world deployment.

Our main contributions are:
\begin{itemize}[leftmargin=1.25em,itemsep=0.15em,topsep=0pt]
    \item \textbf{\ACRO}: an image-space contact-overlay method
    that drops into any image-conditioned policy with no
    architectural changes and no force calibration. The same
    pipeline transfers unmodified across three policy families
    (BC-Transformer, miniVLA, $\pi_{0.5}$).
    \item A low-cost, open-source \textbf{magnetic contact sensor}
    and \textbf{parametric CAD-to-mold pipeline} that generates
    3D-printable molds from user-specified geometry. The sensor
    costs under \$5 in elastomer and magnet parts (under \$50
    assembled) and is validated end-to-end on four real-world
    contact-rich tasks with $\pi_{0.5}$.
    \item A characterization of \textbf{when and how contact overlays
help}: gains scale with task headroom and grasp structure, peaking
on multi-grasp long-horizon tasks; image-space delivery outperforms a compact state-vector baseline and information-matched separate-stream alternatives; and optimal overlay granularity matches the spatial consistency of the underlying contact.
\end{itemize}
\section{Related Work}
\label{sec:related}

\paragraph{Contact and tactile sensing for visuomotor policies.}
A broad literature explores how to make manipulation policies
contact-aware; see~\citet{xie2025forceful} for a recent survey. We
focus on \emph{how} contact signals are integrated into the policy.
Three strategies dominate. The first concatenates contact readings
to the proprioceptive state vector (Lee et al.~\cite{lee2019making};
FACTR~\cite{liu2025factr}, which argues this under-utilizes the
modality). The second adds a separate tactile encoder
(ManiFeel~\cite{xu2025manifeel}, CONTACT~\cite{saka2026contact},
Tactile-VLA~\cite{huang2025tactilevla}), with related variants in
ForceVLA~\cite{yu2026forcevla}, FoAR~\cite{he2024foar}, and
HapticVLA~\cite{he2025hapticvla}. The third grounds tactile signals
geometrically as visuotactile point clouds (Robot
Synesthesia~\cite{yuan2023robotsynesthesia},
3D-ViTac~\cite{huang20243dvitac}, SaTA~\cite{huang2025spatially}),
at the cost of point-cloud backbones incompatible with the 2D
image-token architectures of modern VLAs. Orthogonal to all of
these, CAP~\cite{cui2026contact} reframes contact as task
\emph{specification} rather than sensed feedback. \ACRO\ differs
from each line by introducing contact information into the policy's
existing image input, with no new state dimensions, encoders, or
architectural changes.

\paragraph{Visual overlays as a policy interface.}
A recent line renders auxiliary information directly into the
policy's image input rather than as a separate modality.
RT-Trajectory~\cite{gu2024rttrajectory} overlays trajectory
sketches; TraceVLA~\cite{zheng2024tracevla} overlays past motion
traces; AimBot~\cite{dai2025aimbot} overlays end-effector reticles.
A parallel line in the VLM literature overlays candidate actions,
keypoints, or affordances for vision-language reasoning
(PIVOT~\cite{nasiriany2024pivot}, MOKA~\cite{liu2024moka}).
Audio-VLA~\cite{wei2025audiovla} also augments LIBERO with a
contact-derived signal but routes it through a separate audio
stream. These methods all overlay signals that are kinematically
dense and smooth, or carried in a separate modality. We extend the
paradigm to a qualitatively different signal: contact events that
are sparse, event-driven, and visually hidden at the moments they
matter.

\paragraph{Low-cost tactile sensors.}
Vision-based sensors (GelSight~\cite{yuan2017gelsight},
DIGIT~\cite{lambeta2020digit}) provide rich contact images but
require dedicated optics; piezoelectric arrays capture only normal
contact~\cite{huang20243dvitac}. Magnetic sensors offer robustness
by decoupling sensing electronics from a replaceable contact
interface. ReSkin~\cite{bhirangi2021reskin} and
AnySkin~\cite{bhirangi2024anyskin} embed magnetic microparticles in
elastomers but require specialized magnetization equipment that
limits accessibility. A separate line embeds pre-magnetized
discrete magnets in cast elastomers, yielding durability at the
cost of manually designed, geometry-specific
molds~\cite{uskin,dai2022design,rehan2022soft}.
eFlesh~\cite{pattabiraman2025eflesh} restores parametric
customizability by 3D-printing the entire sensor in TPU but
requires stacked layers (24\,mm in the reference design) and
remains susceptible to lattice delamination. Our sensor closes
this gap with an automated CAD-to-mold pipeline that yields a
4\,mm cast body for under \$5 in elastomer and magnet components.
\section{\ACRO: Sensing and Rendering Contact}
\label{sec:method}

\ACRO\ has two coupled components: (i) a rendering pipeline that
projects per-element contact measurements onto the policy's RGB
observations as a spatial overlay, and (ii) a low-cost magnetic
tactile sensor that drives the pipeline on real hardware. The
rendering pipeline is sensor-agnostic: it operates on raw
per-element measurements without calibration to physical force
units, and accepts any tactile sensor providing 3-axis per-element
readings (or simulated contact forces). 

\subsection{Overlay Representation}
\label{sec:method:overlay}

For an image-conditioned policy, raw tactile readings are hard to
leverage without spatial grounding: a sensor element index alone
does not tell the policy where in the visual field the contact
occurs. We instead project each element's measurement onto the
camera image using the robot's live forward kinematics and
calibrated camera parameters, producing a contact overlay
geometrically registered to the gripper at every timestep. The
same pipeline applies whether the underlying contact source is
simulated forces or raw sensor readings on real hardware.

\paragraph{Sensor model.}
We assume an abstract tactile sensor consisting of $N$ sensing
elements per finger, with nominal positions $\{\mathbf{p}_i\}_{i=1}^{N}
\subset \mathbb{R}^3$ in a sensor local frame $\mathcal{F}_s$. Each
element reports a 3-axis measurement $\mathbf{s}_i \in \mathbb{R}^3$:
a decomposed contact force vector in simulation, or a raw magnetic
flux deflection on real hardware. The two sources are on different
absolute scales but are treated identically by the rendering
pipeline; \ACRO\ does not assume calibrated forces. For a
parallel-jaw gripper, we render overlays separately for each finger
$f \in \{L, R\}$; we denote the measurements at finger $f$ as
$\{\mathbf{s}^{(f)}_i\}_{i=1}^{N}$ and the per-finger aggregate as
$\bar{\mathbf{s}}^{(f)} = \frac{1}{N}\sum_{i=1}^{N}
\mathbf{s}^{(f)}_i$. The pipeline is agnostic to the specific value
of $N$ or the spatial layout of $\{\mathbf{p}_i\}$, accommodating
any parallel-jaw tactile sensor that produces per-element 3-axis
readings.

\paragraph{Projection.}
Each sensing element is rendered as an arrow anchored at its
projected position in the image (see example in \cref{fig:overview}). The arrow tail and tip are
obtained by transforming the local-frame position $\mathbf{p}_i$
(scaled by spatial factor $\sigma > 0$, with the tip additionally
offset by the measurement $\mathbf{s}_i$ scaled by $\lambda > 0$)
through the sensor-to-camera chain using forward kinematics and
known camera extrinsics, then projecting through a standard pinhole
camera model. The wrist-camera extrinsic is updated each timestep
from the robot's joint configuration; the external-camera extrinsic
is calibrated once offline. The resulting screen-space arrow length
reflects both measurement intensity and perspective foreshortening
at the element's depth, providing a depth-consistent visual cue.
The scale $\lambda$ is tuned per sensor source so that typical
contact events produce visually visible arrows; \ACRO\ requires no
further force calibration of the sensor signal beyond this single scalar.
Full derivations are in Appendix~\ref{app:projection}.
Arrows and projected position markers (rendered at the projected
$\mathbf{p}_i$ even when $\mathbf{s}_i = \mathbf{0}$) are then
alpha-blended onto the base RGB image with per-layer opacity
$\alpha \in (0, 1]$. Distinct per-finger colors let the policy
recover finger identity from color alone. The augmented image
preserves the resolution, channel count, and pixel range of the
original observation, serving as a drop-in replacement for the
policy's RGB input.

\paragraph{Overlay variants.}
We study three variants spanning a granularity spectrum.
\textbf{Multi-arrows} renders one arrow per sensing element using
$\mathbf{s}^{(f)}_i$ as the arrow's direction and magnitude,
preserving spatial structure at the cost of visual clutter.
\textbf{Avg.\ Arrow} renders a single per-finger arrow at the
projected finger center, using the per-finger aggregate
$\bar{\mathbf{s}}^{(f)}$ as the arrow's direction and magnitude.
\textbf{Binbars} discards both location and direction, rendering a
vertical bar of height proportional to $\|\bar{\mathbf{s}}^{(f)}\|$
at a fixed image location per finger, isolating whether the policy
uses the overlay as a directional cue or as a contact-event signal. Examples of each are shown in \cref{fig:real_overlays}.
We compare all three in both simulated and real robot experiments (\cref{sec:experiments}).

\subsection{Custom Magnetic Tactile Sensor}
\label{sec:hardware}

\begin{figure}[b]
    \centering
    \includegraphics[width=0.99\linewidth]{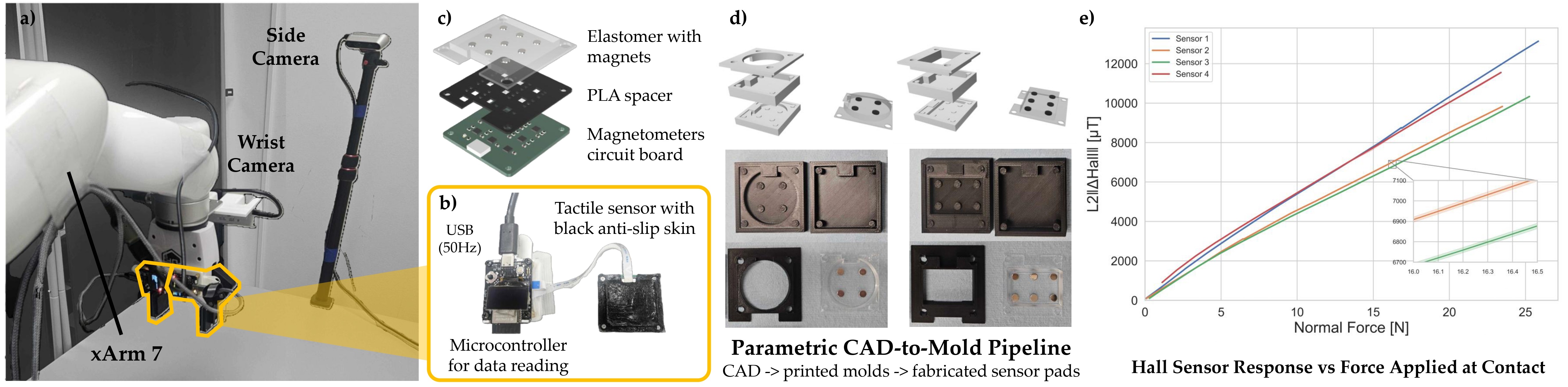}
    \caption{
    \footnotesize{
    \textbf{Hardware system.}
    \textbf{(a)}~Visuo-tactile setup: UFactory xArm7 with two
    RealSense cameras and two tactile sensors mounted on the
    gripper fingers.
    \textbf{(b)}~Custom tactile sensor with anti-slip skin.
    \textbf{(c)}~Internal structure of the tactile sensor.
    \textbf{(d)}~Our parametric CAD-to-mold pipeline: CAD-generated
    alternative casting molds (top) and the 3D-printed molds with
    the fabricated sensor pads (bottom).
    \textbf{(e)}~Physical characteristics of the sensor.}
    }
    \label{fig:hardware_setup}
    \vspace{-0.25cm}
\end{figure}

The rendering pipeline can be driven by any sensor providing
per-element 3-axis readings. For our real-world experiments, we
build a low-cost magnetic tactile sensor with a specific instance
of $N{=}9$ sensing elements arranged in a $3{\times}3$ grid per
finger (\cref{fig:hardware_setup}). The sensor captures contact
through 3D magnetic flux: external forces deform a soft elastomer
surface embedded with permanent magnets, and the resulting field
changes are recorded by an underlying magnetometer array. The
sensor maintains a slim 4\,mm profile.
Each sensor pairs a soft silicone elastomer pad (XP-565) embedded
with a $3{\times}3$ array of permanent magnets and a custom PCB
housing a matching array of 3-axis magnetometers (Allegro A31031),
separated by a 3D-printed PLA spacer that offsets surface-mount
components (\cref{fig:hardware_setup}c). An ESP32 microcontroller
adapted from~\cite{tanaka2025mechanisms} polls the array via
I$^2$C and streams raw flux at 50\,Hz. The modular construction
decouples sensing electronics from the contact interface, allowing
the elastomer layer to be replaced when worn.

A parametric CAD-to-mold pipeline programmatically generates
multi-stage 3D-printable casting molds from dimension and
magnet-layout inputs (\cref{fig:hardware_setup}d), accommodating
sensor geometries beyond the specific $3{\times}3$ instance used
in our experiments. The elastomer is cast in a base mold,
semi-cured ($\sim$40\% of curing time), then transferred to a
secondary mold where magnets are seated and encapsulated by a top
pour. After curing, the elastomer base is bonded to the PLA spacer
(Loctite SF770 primer, Loctite 406 cyanoacrylate), an anti-slip
skin is applied, and the assembly is fastened to the PCB and
gripper with machine screws. 

The aggregate magnetic flux across the sensor array varies
monotonically with applied normal force
(\cref{fig:hardware_setup}e; 4 independently fabricated sensors of the same $3{\times}3$ layout). \ACRO\ uses these raw readings directly; the sensor costs under \$5 in parts (excluding the PCB) or under \$50 assembled, and is open-sourced. Details are provided in Appendix~\ref{app:hardware}.
\section{Experiments}
\label{sec:experiments}


\begin{figure}[b]
    \centering
    \includegraphics[width=\linewidth]{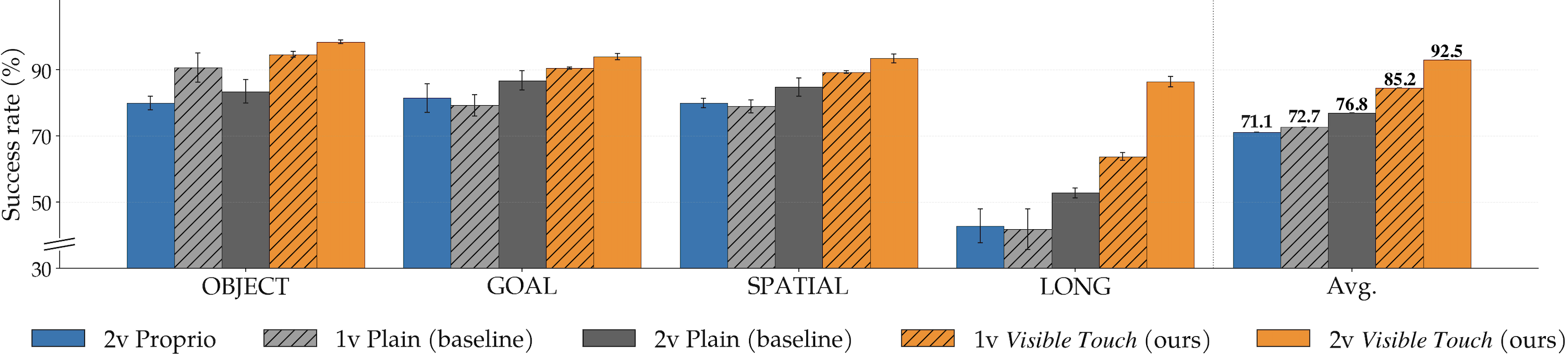}
    \caption{
    \small{\textbf{BC-Transformer on LIBERO across camera
    configurations} ($n{=}3$ seeds, error bars are std).\
    \textcolor{orange}{Orange} bars show \ACRO\ (ours);
    \textcolor{gray}{gray} bars show plain baselines;
    \textcolor{blue}{blue} bars show the Proprio baseline that
    receives an aggregated 12-D contact wrench through the proprioceptive state. \ACRO\
    consistently improves over plain images in both 1-view and
    2-view, with the largest gain on LIBERO\_LONG ($+33.7$pp
    2-view). See details in \cref{app:bct}.
    }}
    \label{fig:bct-main}
\end{figure}

We evaluate \ACRO\ on both a controlled simulation benchmark and real-world manipulation tasks, across three policy
architectures (BC-Transformer from
LIBERO~\cite{liu2023libero}, miniVLA~\cite{belkhale2024minivla},
and $\pi_{0.5}$~\cite{black2025pi05}). We aim to isolate the effect
of overlay rendering from confounds of architecture, training data,
or sensor noise, and characterize the conditions under which
overlays provide the largest benefit. Concretely, we design our
experiments to answer three research questions:

\begin{itemize}[leftmargin=1.75em,itemsep=0.02em,topsep=0.1em]
\item[\textbf{Q1.}] \textbf{Does \ACRO\ improve visuomotor policy
performance, and does the improvement transfer across architectures?}
We evaluate on the LIBERO benchmark with two policies trained or
fine-tuned in simulation, and on four real-world contact-rich tasks
with $\pi_{0.5}$.
\item[\textbf{Q2.}] \textbf{How does image-space contact rendering compare with alternative contact-integration strategies?} We compare Visible Touch against a compact state-vector baseline in simulation and an information-matched separate-image-stream baseline in the real world.
\item[\textbf{Q3.}] \textbf{When and how do overlays help most?}
We analyze the per-task structure of overlay gains, and ablate
overlay granularity in simulation and real-world.
\end{itemize}


\subsection{Simulation Experiments on LIBERO}
\label{sec:exp:libero}

We evaluate \ACRO\ on the LIBERO simulation benchmark with two
policies: a from-scratch \textbf{BC-Transformer} for controlled
input-condition comparisons, and a pretrained \textbf{miniVLA} that
we both train from scratch (VLM weights) and fine-tune with contact overlays to
test transfer to VLAs.

\paragraph{Setup.}
We use all LIBERO suites~\cite{liu2023libero}. Since LIBERO does
not expose native tactile observations, we extract per-contact
force vectors from the MuJoCo contact array (filtered to
gripper-finger geometries) as the per-element measurements
$\mathbf{s}_i$ for overlay rendering. For the
BC-Transformer~\cite{liu2023libero}, we train in the
\texttt{multitask} configuration with 3 seeds, evaluating
final-epoch checkpoints with 20 rollouts per task (see \cref{app:bct}). We compare
three input conditions: \textbf{Plain} (LIBERO's default RGB with
8-D proprioception); \textbf{Proprio} (Plain plus a 12-D contact
wrench concatenated to the state); and \textbf{\ACRO} (Plain plus
our image-space overlay; default Avg.\ Arrow). The Proprio baseline provides an aggregated 12-D contact wrench through the state vector, whereas \ACRO\ renders contact in image space; this comparison evaluates image-space rendering against a compact state-vector integration rather than holding the representation and information content exactly constant. We test both 1-view
(\texttt{agentview}) and 2-view (\texttt{agentview} + wrist)
configurations. For miniVLA~\cite{belkhale2024minivla} (Qwen2.5
0.5B backbone, DINOv2 + SigLIP encoders at $224$\,px), we
pretrain from scratch on LIBERO\_90 (vs.\ the publicly released
\texttt{minivla-libero90-prismatic} baseline) and separately
fine-tune the pretrained model on each suite with and without
overlays (see \cref{app:minivla_libero}). 

\begin{table}[t]
\small
\centering
\caption{\footnotesize{
miniVLA success rates (\%) on LIBERO across pretraining and fine-tuning.
LIBERO\_90 pretraining results are from single runs. For fine-tuning,
baseline entries are the strongest single-run baselines available per
suite, while \ACRO\ results are mean $\pm$ s.e. over 5 training seeds
using the suite-specific headline recipe.
}}
\label{tab:minivla_finetuning}
\renewcommand{\arraystretch}{1.2}
\begin{tabularx}{\linewidth}{l Y YYYY Y}
\toprule
& \textbf{Pretraining} & \multicolumn{5}{c}{\textbf{Fine-tuning}}\\
\cmidrule(lr){2-2} \cmidrule(lr){3-7}
& LIBERO\_90 & OBJECT & GOAL & SPATIAL & LONG & \textbf{Avg.} \\
\midrule
Baseline
& 48.3\% 
& 51.0\%
& 48.5\%
& 71.5\%
& 32.5\%
& 50.9\% \\

\ACRO
& \textbf{63.2\%} 
& $\mathbf{70.6 \pm 4.6}$\%
& $\mathbf{80.5 \pm 4.3}$\%
& $\mathbf{87.0 \pm 1.8}$\%
& $\mathbf{66.6 \pm 2.2}$\%
& \textbf{76.2\%} \\
\bottomrule
\end{tabularx}
\vspace{-1em}
\end{table}

\paragraph{Q1: Overlays consistently improve performance and
transfer across architectures.}
For BC-Transformer (\cref{fig:bct-main}), \ACRO\ outperforms the
plain baseline in every suite and both camera configurations. In
2-view, \ACRO\ adds $+15.7$pp on average ($+33.7$pp on LIBERO\_LONG);
1-view shows the same pattern ($+12.5$pp average), confirming the
benefit does not require a wrist camera. The wrist and the overlay
are complementary rather than substitutable: adding the wrist alone
yields modest gains ($+4.1$pp average, $-7.9$pp on OBJECT), while
wrist-plus-overlay is best on every suite. The same pattern
transfers to miniVLA (\cref{tab:minivla_finetuning}): Our overlay-pretrained miniVLA reaches 63.2\% on LIBERO-90 versus 48.3\% for the released Stanford checkpoint evaluated in the matched single-image, no-chunking configuration (+14.9pp, 90 tasks × 20 trials), and
fine-tuning yields a $+25.3$pp average gain ($+34.1$pp on
LIBERO\_LONG). See details in~\cref{app:minivla_libero}.


\paragraph{Q2: Image-space delivery outperforms a compact state-vector baseline and an information-matched separate-stream alternative.}
We compare \ACRO\ against the Proprio baseline, which provides an aggregated 12-D contact wrench concatenated to the proprioceptive state. In the 2-view setting, \ACRO\ exceeds Proprio by $+21.4$pp on average, with the largest gap on LIBERO\_LONG ($+43.7$pp). Notably, Proprio underperforms the plain 2-view baseline on every suite ($-5.7$pp on average), suggesting that direct concatenation of contact information to the state vector is a relatively weak integration strategy in this setup. In contrast, image-space rendering yields consistent improvements, demonstrating that \ACRO\ is more effective than this compact state-vector baseline for BC-Transformer.


\begin{table}[t]
\small
\centering
\caption{\footnotesize{Success rates (\%) for overlay variants on
LIBERO suites (BC-Transformer, 2-view, $n{=}3$ seeds).}}
\vspace{0.1cm}
\label{tab:libero_overlay_ablation}
\renewcommand{\arraystretch}{1.15}
\begin{tabularx}{\linewidth}{l *{5}{Y}}
\toprule
\textbf{\ACRO\ variant} & \textbf{OBJECT} & \textbf{GOAL} & \textbf{SPATIAL} & \textbf{LONG} & \textbf{Avg.} \\
\midrule
Multi-arrows          & $94.3 \pm 5.6$          & $91.8 \pm 0.5$          & $90.2 \pm 1.0$          & $76.2 \pm 2.7$          & $88.1$ \\
Avg.\ Arrow  & $\mathbf{98.2 \pm 1.2}$ & $\mathbf{92.8 \pm 2.5}$ & $92.3 \pm 2.7$          & $\mathbf{86.5 \pm 2.3}$ & $\mathbf{92.5}$ \\
Binbars               & $98.0 \pm 0.9$          & $92.7 \pm 1.7$          & $\mathbf{94.5 \pm 1.5}$ & $78.8 \pm 2.6$          & $91.0$ \\
\bottomrule
\end{tabularx}
\vspace{-1.25em}
\end{table}

\begin{figure}
    \centering
    \begin{subfigure}[b]{0.5\linewidth}
        \centering
        \includegraphics[width=\linewidth]{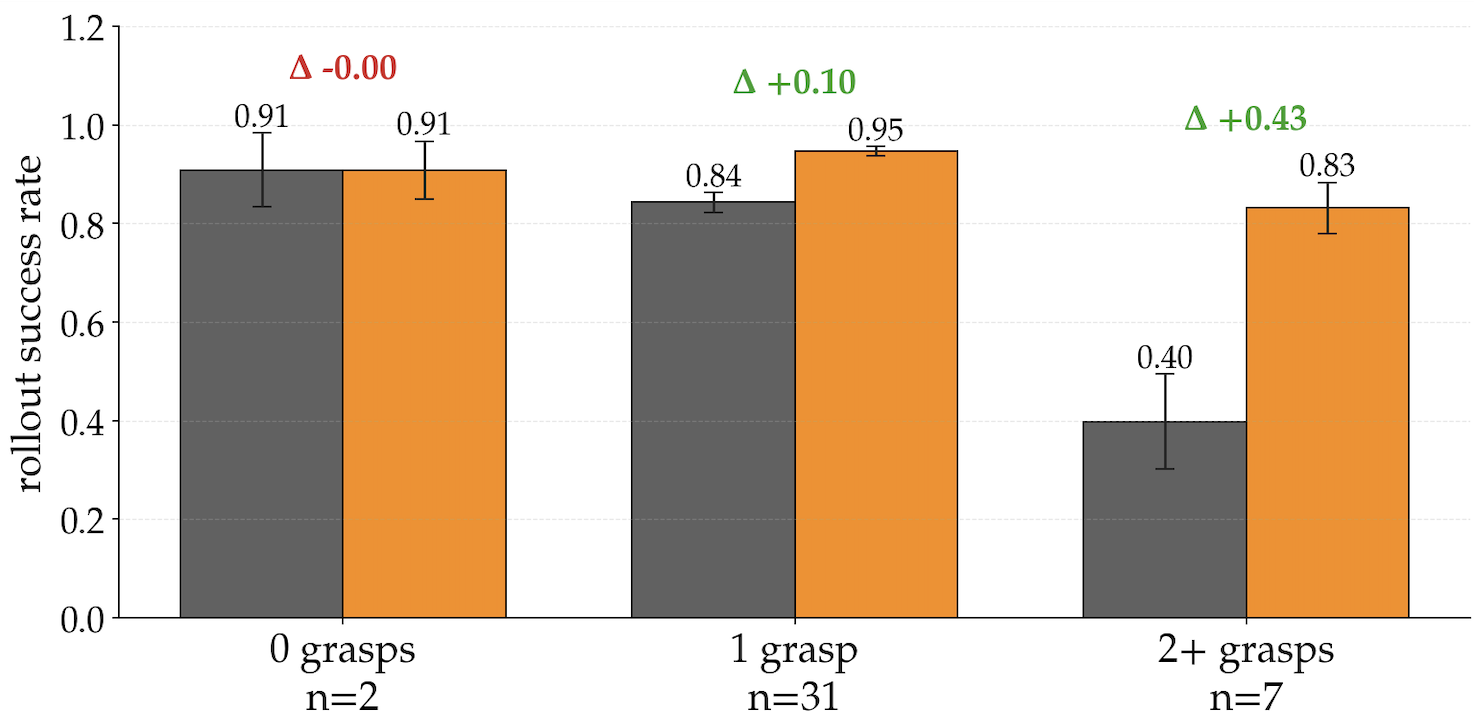}
        \caption{BC-Transformer on LIBERO suites (40 tasks).}
        \label{fig:grasp_buckets_bct}
    \end{subfigure}
    \hfill
    \begin{subfigure}[b]{0.48\linewidth}
        \centering
        \includegraphics[width=\linewidth]{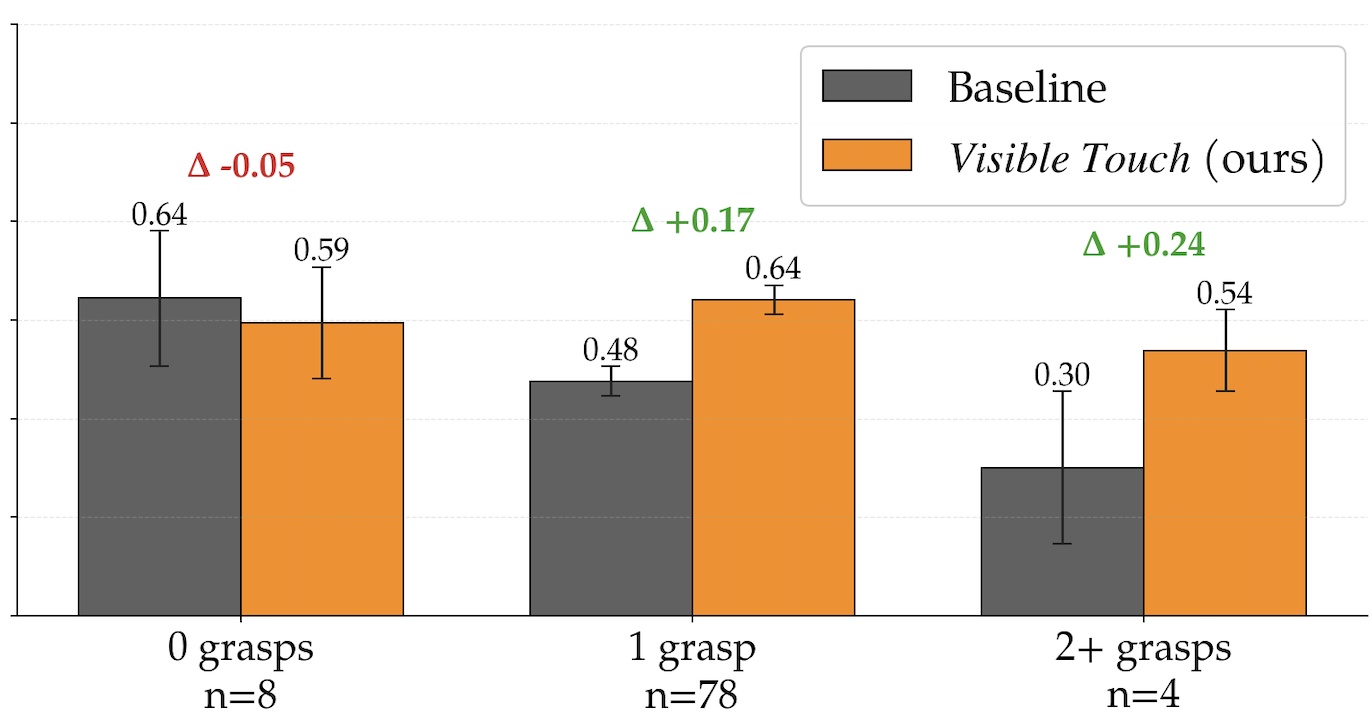}
        \caption{miniVLA on LIBERO\_90 (90 tasks).}
        \label{fig:grasp_buckets_minivla}
    \end{subfigure}
    \caption{Success rates of \ACRO\ (Avg.\ Arrow overlay) over
    plain baselines, bucketed by gripper-close events. Tasks with
    no grasps show no benefit; single-grasp tasks show modest
    gains; multi-grasp tasks show the largest gains. The pattern
    holds across both BC-Transformer
    (\subref{fig:grasp_buckets_bct}) and a pretrained miniVLA on
    LIBERO\_90 (\subref{fig:grasp_buckets_minivla}).}
    \label{fig:libero_grasp_buckets}
  
\end{figure}

\paragraph{Q3a: Overlay benefit scales with task headroom and grasp
structure (across architectures).}
For BC-Transformer, the per-task improvement
$\Delta = \text{\ACRO} - \text{plain}$ across all 40 LIBERO tasks
(2-view) correlates strongly with task headroom
($r = 0.924$, $p < 10^{-3}$): overlays help most where vision-only
performance is lowest. Bucketing tasks by gripper-close events
(\cref{fig:grasp_buckets_bct}) reveals the same pattern through
task structure: 
multi-grasp tasks ($n=7$) show
$\Delta = +43.3$pp. 
After controlling for headroom via partial
correlation, two task properties retain independent predictive
power for $\Delta$: contact-event count
($r = -0.571$, $p < 10^{-3}$) and trajectory length
($r = -0.368$, $p = 0.019$).
The grasp-structure pattern replicates on miniVLA
(\cref{fig:grasp_buckets_minivla}): across the 90 LIBERO\_90 tasks,
 1-grasp tasks ($n=78$) show $+17$pp, and 2+-grasp tasks ($n=4$) show $+23.8$pp. 
Grasp structure governs overlay benefit consistently across architectures.

\paragraph{Q3b: Overlay granularity has a non-monotonic optimum.}
\cref{tab:libero_overlay_ablation} compares three overlay variants
spanning a granularity spectrum. The coarser Binbars representation
is competitive with the default Avg.\ Arrow, matching it within
noise on three suites and falling behind only on LIBERO\_LONG
($-7.7$pp). The finer Multi-arrows representation underperforms
Avg.\ Arrow, with the largest deficit on
LIBERO\_LONG ($-10.3$pp). An intermediate level of detail
(per-finger averaging) outperforms both coarser bars and finer
per-contact-point arrows, suggesting that excess spatial detail
can introduce visual clutter that displaces the encoder's attention
from task-relevant features. We return to this finding in the
real-world experiments (Section~\ref{sec:exp:realworld}), where
the optimal granularity reverses.

\subsection{Real-World Validation with $\boldsymbol{\pi_{0.5}}$ }
\label{sec:exp:realworld}

\begin{figure}[t]
    \centering
    \begin{subfigure}[t]{0.565\linewidth}
        \centering
        \includegraphics[width=\linewidth]{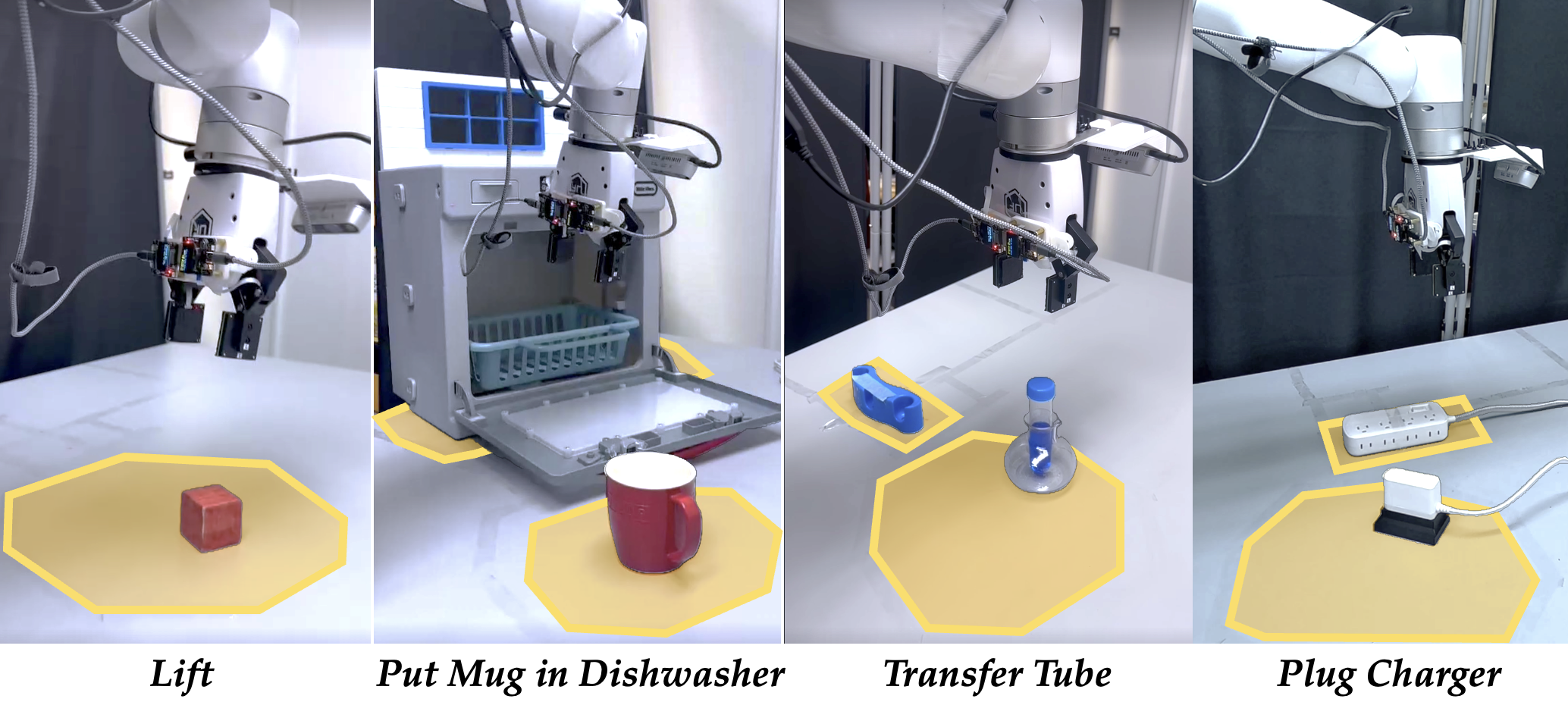}
        \caption{Real-world tasks (yellow areas highlight initial
        state randomization).}
        \label{fig:realworld_tasks}
    \end{subfigure}
    \hfill
    \begin{subfigure}[t]{0.428\linewidth}
        \centering
        \includegraphics[width=\linewidth]{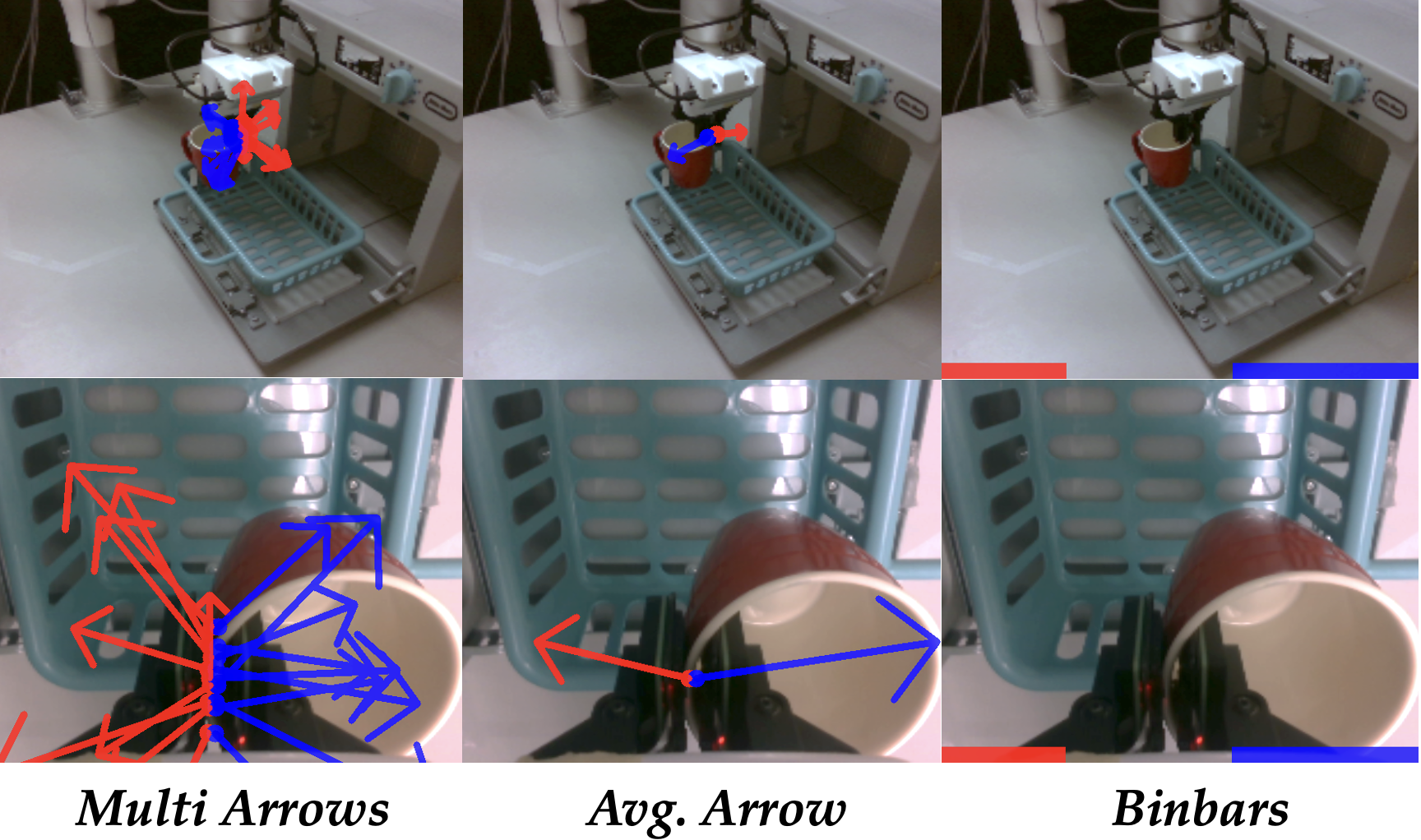}
        \caption{Variants of \ACRO contact overlay.}
        \label{fig:real_overlays}
    \end{subfigure}
    \caption{\textbf{Real-world experimental setup.}
    \textbf{(a)}~Contact-rich manipulation tasks: \textit{Lift, Transfer
    Tube, Put Mug in Dishwasher}, and \textit{Plug Charger}.
    \textbf{(b)}~Example contact overlays rendered onto external and
    wrist camera views.}
    \label{fig:realworld_setup}
    \vspace{-0.25cm}
\end{figure}

\begin{wrapfigure}{r}{0.26\linewidth}
    \centering
    \vspace{-1.5em}
    \includegraphics[width=\linewidth]{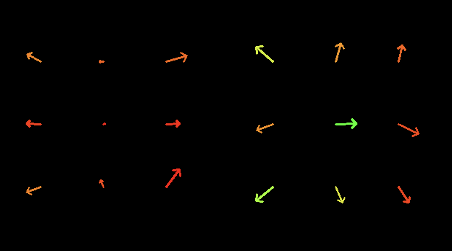}
    \caption{\small \textbf{Tac-View}: contact arrows
     as a separate image.}
    \label{fig:tac_view}
    \vspace{-1.25em}
\end{wrapfigure}
\paragraph{Setup.}
We fine-tune $\pi_{0.5}$~\cite{black2025pi05} (\texttt{pi05\_droid}
checkpoint, 2.3B) with
LoRA~\cite{hu2021loralowrankadaptationlarge} on two image views
(full details in \cref{app:impl:pi05}). We compare eight conditions:
\textbf{Baseline} (standard fine-tuning);
\textbf{Tac-View}, a non-overlay tactile baseline that renders the
same contact arrows as a separate image stream (\cref{fig:tac_view}) into one of $\pi_{0.5}$'s camera slots, mirroring how vision-based tactile
sensors (e.g., GelSight~\cite{yuan2017gelsight}) are typically
integrated into image-conditioned policies; \textbf{Position-Only}, which renders the projected markers without contact information; two binary-contact ablations, \textbf{Binary Contact (1x)} and \textbf{Binary Contact (9x)}, which preserve contact occurrence at one aggregate location per finger or independently at each taxel location, respectively; and three \ACRO\
overlay variants (Binbars, Avg.\ Arrow, Multi-arrows) that
composite onto the existing RGB views. We evaluate on four
contact-rich manipulation tasks (\cref{fig:realworld_tasks}):
\textbf{Lift} (single-stage: pick up a cube),
\textbf{Transfer Tube} (two-stage: pick up a test tube, insert into
slot), \textbf{Put Mug in Dishwasher} (three-stage: pull rack open,
pick mug, place in rack), and \textbf{Plug Charger} (two-stage:
pick charger, insert into outlet). Each task is scored at the
sub-task level with 30 trials per condition. For aggregate real-world results, we macro-average across tasks: we first average sub-task success within each task, then average the Multi-arrows–Baseline difference across the four tasks. Under this metric, Multi-arrows improves over Baseline by +30.3 pp. Our workspace consists of a UFactory xArm7 with a parallel-jaw gripper, two Intel
RealSense D435 cameras (wrist-mounted and external), and our custom
$3 \times 3$ magnetic contact sensors. We
collect 100 teleoperated demonstrations per task using
TeleDex~\cite{rayyan2026teledexaccessibledexterousteleoperation},
logging all raw sensor streams and applying overlay variants
\emph{post-hoc} so that all conditions are trained on the same
underlying data (\cref{app:realworld}).

\begin{figure}[t]
\centering
\includegraphics[width=\linewidth]{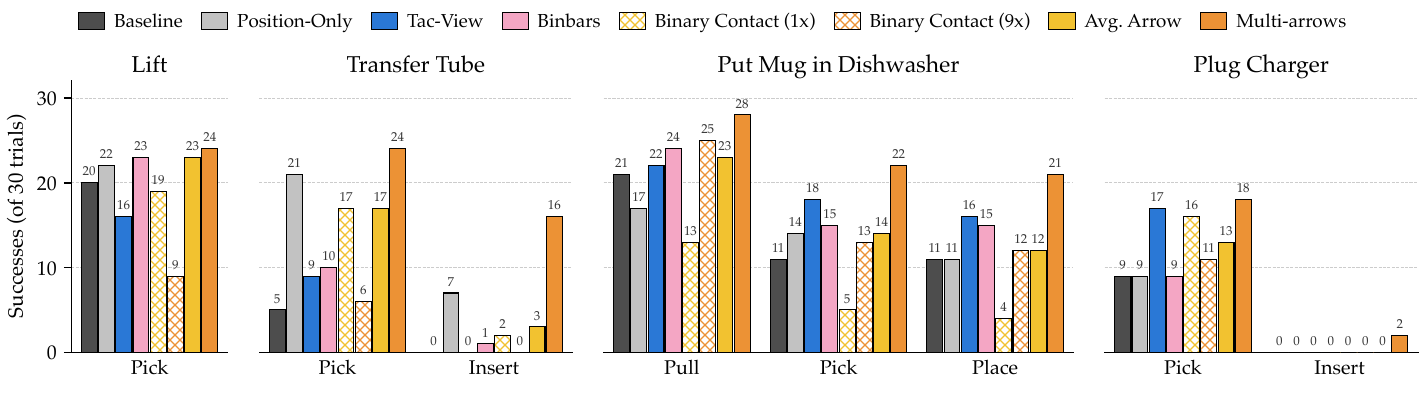}
\caption{Real-world evaluation of $\pi_{0.5}$ fine-tuning across
four contact-rich manipulation tasks (successes out of 30 trials
per condition, sub-task scored). Final-stage success corresponds to
complete end-to-end task success. Hatched bars denote the binary
ablations of the matching solid variant: Binary Contact (1x) and
(9x) keep only per-finger and per-taxel contact occurrence,
respectively. Exact counts are listed in
\cref{tab:realworld_main} (\cref{app:realworld:eval}).}
\label{fig:realworld_main}
\vspace{-1.25em}
\end{figure}

\paragraph{Q1: \ACRO\ transfers to real-world contact-rich
manipulation.}
\ACRO\ improves over the Baseline on nearly every sub-task, with
Multi-arrows strictly outperforming Baseline everywhere
(\cref{fig:realworld_main}). The largest gains appear on
multi-stage tasks where contact information is most diagnostic: on
Transfer Tube, Multi-arrows reaches 24/30 on Pick (vs.\ 5/30) and
16/30 on Insert (vs.\ 0/30); on Put Mug in Dishwasher, the three
stages improve by 7, 11, and 10 trials respectively; on Plug
Charger (the hardest task), Multi-arrows is the only condition
that completes any Inserts (2/30) and doubles Baseline success on
Pick (18/30 vs.\ 9/30). The simpler Lift task shows a more modest
gain (24/30 vs.\ 20/30), consistent with the LIBERO observation
that overlays provide the largest benefit on contact-rich
multi-stage tasks.

\paragraph{Component ablations further isolate the source of these gains. }
Position-Only improves over the plain baseline on several stages, most notably Transfer Tube, indicating that geometrically registered visual markers themselves can provide useful end-effector localization. However, binary contact representations are substantially weaker than their full-vector counterparts on the more challenging real-world tasks: Avg. Arrow generally outperforms Binary Contact (1x), while Multi-arrows substantially outperforms Binary Contact (9x). Together, these results suggest that localization and contact occurrence explain part of the benefit, but preserving contact direction and magnitude becomes important in contact-rich real-world manipulation.

\paragraph{Encoder-architecture robustness.}
Notably, \ACRO's overlay benefit holds across both convolutional
(BC-Transformer's ResNet-18~\cite{he2016deep}) and attention-based
(miniVLA's DINOv2 + SigLIP ViT~\cite{oquab2023dinov2,zhai2023sigmoid}
and $\pi_{0.5}$'s SigLIP-So400m) vision encoders. This suggests
the overlay's effectiveness is not contingent on specific
feature-extraction mechanics, but rather on rendering contact at a
geometrically meaningful image location that the encoder can
attend to regardless of its architectural priors. We leave a
controlled encoder-family comparison (e.g., the same policy
backbone with ResNet vs.\ ViT image encoders) to future work.

\paragraph{Q2: Image-space overlay outperforms separate-stream
tactile rendering.}
Tac-View, which provides the same per-taxel contact arrows as Multi-arrows through a separate image stream rather than compositing them onto the RGB views, improves over the Baseline on three of four tasks (notably +7 on Dishwasher Pick and +8 on Plug Pick), confirming that the contact information itself is useful. However, Multi-arrows
strictly outperforms Tac-View on every sub-task, with the largest
gaps on the Transfer Tube stages (Pick 24/30 vs.\ 9/30, Insert
16/30 vs.\ 0/30). The coarser \ACRO\ variants (Binbars and
Avg.\ Arrow), which discard the per-taxel spatial structure that
both Multi-arrows and Tac-View preserve, win against Tac-View on
the simpler Lift and Tube sub-tasks but lose to it on the more
spatially complex Dishwasher Pick, Dishwasher Place, and Plug Pick
sub-tasks. This mirrors the BC-Transformer Q2 finding
(\cref{sec:exp:libero}) under a different alternative
representation: when contact information is held constant,
rendering it as a spatial overlay on the scene the policy already
attends to can be more effective than supplying it through a
parallel visual channel, provided the overlay preserves enough
spatial detail to carry the underlying signal.

\paragraph{Q3b: Optimal granularity reverses in real-world.}
Multi-arrows, which renders all 9 per-taxel readings per finger,
is strongest in every column. This reverses the simulation finding,
where Avg.\ Arrow dominated Multi-arrows. Importantly, the Binary Contact (9x) ablation shows that finer spatial granularity alone is insufficient: preserving per-taxel contact occurrence without direction or magnitude performs substantially worse than Multi-arrows. Similarly, Avg. Arrow generally outperforms Binary Contact (1x), indicating that contact direction and magnitude provide useful information beyond binary occurrence. We attribute the reversal to a difference in the spatial consistency of the contact signal: in simulation, MuJoCo contacts arise anywhere on the finger
collision mesh, so per-contact arrows occupy inconsistent image
locations across timesteps; on real hardware, signals arrive at
fixed taxel positions on the gripper's inner surface, and the
per-taxel rendering carries spatially stable information that the
per-finger average discards.

\section{Conclusion}
\label{sec:conclusion}

We introduce \ACRO, a method that renders contact information as
image-space overlays on the policy's RGB observations, requiring no
architectural changes. Across LIBERO, image-space overlays improve
BC-Transformer success over plain images and a
compact proprioceptive baseline. The pattern transfers across
architectures: miniVLA gains 25 percentage points on LIBERO fine-tuning, and
$\pi_{0.5}$ gains 30 percentage points across four real-world contact-rich tasks. We hope \ACRO\ provides a
practical path for adding contact awareness to image-conditioned
policies, including pretrained VLAs.

\paragraph{Limitations.}
Our proprio baseline tests one specific state-vector integration;
alternative implementations (e.g., learned tactile encoders, separate
input streams) may close some of the gap, though we expect the
direction of the finding to hold. Our real-world evaluation spans
four tasks on a single robot with a parallel-jaw gripper;
generalization to other morphologies, deformable objects, and
unstructured settings remains future work. Our magnetic sensor is
tested at a single $3\times 3$ geometry, and we have not yet
demonstrated cross-geometry portability of trained policies. Finally,
the conjecture that optimal overlay granularity tracks the spatial
structure of the contact signal merits further investigation with
finer-grained sensors and additional simulators.

\clearpage
\acknowledgments{This work was conducted at the University of California, Los Angeles, with institutional and computational support from the UCLA Department of Computer Science. We thank the members of the UCLA Robot Intelligence Laboratory and RoMeLa for helpful discussions and support throughout the project.}


\bibliography{reference}  

\clearpage
\appendix
\crefalias{section}{appendix}

\addcontentsline{toc}{section}{Appendix} 
\part{Appendix} 
\parttoc 

\newpage

\section{Projection Pipeline Derivation}
\label{app:projection}

This appendix provides the full math for the projection pipeline
described in Section~\ref{sec:method:overlay}.

\paragraph{Kinematic chain.}
Let $\mathbf{q} \in \mathbb{R}^7$ be the current joint
configuration. The rigid-body transform from robot base frame to sensor frame is
\begin{equation}
  {}^{b}\mathbf{T}_{s}(\mathbf{q}) = \mathrm{FK}(\mathbf{q}) \cdot {}^{f}\mathbf{T}_{s},
\end{equation}
where $\mathrm{FK}(\mathbf{q}) \in SE(3)$ is the fingertip pose
from forward kinematics and ${}^{f}\mathbf{T}_{s} \in SE(3)$ is
the fixed finger-to-sensor transformation.

For the external camera, the extrinsic ${}^{c}\mathbf{T}_{b} \in
SE(3)$ is calibrated once offline. For the wrist camera, it is
updated each timestep:
\begin{equation}
  {}^{c}\mathbf{T}_{b}(\mathbf{q}) = {}^{c}\mathbf{T}_{\mathrm{wrist}} \cdot \mathrm{FK}_{\mathrm{wrist}}(\mathbf{q})^{-1},
\end{equation}
where ${}^{c}\mathbf{T}_{\mathrm{wrist}} \in SE(3)$ is the fixed
camera-to-wrist calibration and
$\mathrm{FK}_{\mathrm{wrist}}(\mathbf{q}) \in SE(3)$ is the wrist
pose from forward kinematics.

\paragraph{Arrow tail and tip in camera frame.}
Each sensing element is rendered as an arrow with spatial scale
$\sigma > 0$ controlling the apparent sensor footprint and
$\lambda > 0$ controlling the arrow length per unit measurement.
The 3D tail and tip of element $i$'s arrow in camera frame are
\begin{equation}
  \mathbf{X}^{\mathrm{tail/tip}}_i = {}^{c}\mathbf{T}_{b} \cdot {}^{b}\mathbf{T}_{s}(\mathbf{q}) \cdot
  \begin{bmatrix}\sigma\,\mathbf{p}_i + \mu\,\lambda\,\mathbf{s}_i \\ 1\end{bmatrix},
  \quad \mu \in \{0, 1\},
\end{equation}
where $\mu = 0$ yields the tail and $\mu = 1$ yields the tip.
Positions and displacements are defined in the sensor local frame
before any rigid-body transform, so the rendered arrow direction
reflects the sensor-frame axes as seen from the camera.

\paragraph{Pinhole projection.}
Applying the standard pinhole projection
\begin{equation}
  \pi(\mathbf{X}) = \left(\frac{f_x X}{Z} + c_x,\, \frac{f_y Y}{Z} + c_y\right)^\top
\end{equation}
to $\mathbf{X}^{\mathrm{tail/tip}}_i$ yields the tail and tip pixels
$\mathbf{u}_i^{\mathrm{tail}}, \mathbf{u}_i^{\mathrm{tip}} \in
\mathbb{R}^2$, where $(f_x, f_y, c_x, c_y)$ are the camera
intrinsics. The screen-space arrow length
$\|\mathbf{u}_i^{\mathrm{tip}} - \mathbf{u}_i^{\mathrm{tail}}\|$
reflects both measurement intensity and perspective foreshortening
at the element's depth, providing a depth-consistent visual cue at
no additional cost.

\paragraph{Per-variant rendering.}
For the Multi-arrows variant, the procedure above is applied
independently to each of the $N$ sensing elements per finger. For
the Avg.\ Arrow variant, the per-element measurements
$\{\mathbf{s}_i\}_{i=1}^{N}$ are aggregated to a single per-finger
vector before rendering at the projected finger center. For the
Binbars variant, the per-finger aggregate magnitude
$\|\mathbf{s}_f\|$ drives a vertical bar at a fixed image location.
All overlays use high-contrast per-finger colors against the scene
background, with left finger rendered cyan/blue and right finger rendered
magenta/red for finger-identity recovery from color alone. In Tac-View, the arrow colors represent the normalized $z$-readings of the tactile sensor, mapped to a scale where green corresponds to 0 and red corresponds to 1.

\newpage
\section{Tactile Sensor Hardware Details}
\label{app:hardware}

\subsection{Bill of Materials}
\label{app:hardware:bom}

Table~\ref{tab:bom} lists the per-sensor material cost for the
$3{\times}3$ configuration used in our experiments. Costs are
separated into the soft sensor pad (consumable, replaceable) and the
PCB electronics (reusable). All prices are approximate and reflect
small-quantity orders; PCB costs in particular are heavily
order-quantity dependent.

\begin{table}[h]
\centering
\small
\caption{Bill of materials for one tactile sensor ($3{\times}3$
configuration). \textbf{Sensor pad} components are consumable and
can be replaced without replacing the PCB. \textbf{PCB} costs are
rough estimates and vary significantly with order quantity.}
\label{tab:bom}
\setlength{\tabcolsep}{5pt}
\begin{tabular}{lcccc}
\toprule
\textbf{Component} & \textbf{Spec} & \textbf{Qty} & \textbf{Unit price} & \textbf{Subtotal} \\
\midrule
\multicolumn{5}{l}{\textit{Sensor pad (consumable)}} \\
XP-565 silicone elastomer    & 10:1 base/activator & 10\,g  & \$0.05/g  & \$0.50 \\
Neodymium magnets            & $\varnothing$2\,mm $\times$ 1\,mm & 9    & \$0.07/ea & \$0.63 \\
PLA filament (molds + spacer)& ---                 & 20\,g  & \$0.013/g & \$0.26 \\
Loctite SF\,770              & ---   & 1\,ml  & \$0.50/ml & \$0.50 \\
Loctite 406                  & ---       & 0.5\,ml & \$1.65/ml & \$0.83 \\
Anti-slip skin               & ---        & 1 ml & \$0.10/ml    & \$0.10 \\
Machine screws               & M2$\times$6                  & 4      & \$0.20/ea & \$0.80 \\
\midrule
\multicolumn{4}{r}{\textbf{Sensor pad subtotal}} & \textbf{\$3.62} \\
\midrule
\multicolumn{5}{l}{\textit{PCB electronics (reusable)}} \\
Magnetometer array PCB       & $3{\times}3$ Allegro A31031 & 1 & $\sim$\$20 & $\sim$\$20 \\
Microcontroller board        & ESP32-based~\cite{tanaka2025mechanisms} & 1 & $\sim$\$20 & $\sim$\$20 \\
Flexible flat cable       & 10 pin 0.5 mm, 15 mm length& 1 & \$1/ea & \$1 \\
\midrule
\multicolumn{4}{r}{\textbf{Electronics subtotal}} & $\sim$\textbf{\$41} \\
\midrule
\multicolumn{4}{r}{\textbf{Total (fully assembled)}} & $<$\textbf{\$50} \\
\bottomrule
\end{tabular}
\end{table}

\subsection{Fabrication Procedure}
\label{app:hardware:fab}

\begin{figure}[h]
    \centering
    \includegraphics[width=\linewidth]{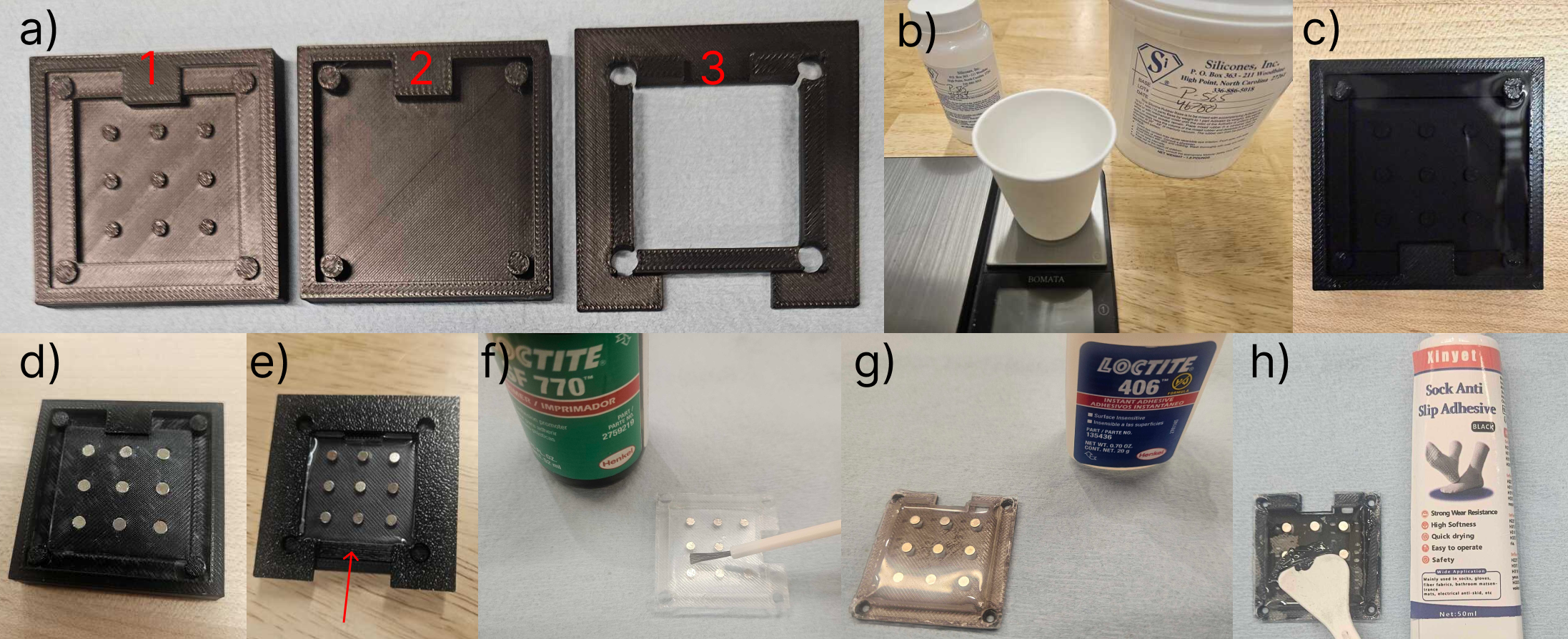}
    \caption{\small Fabrication steps of our custom tactile sensor.}
    \label{fig:appendix_fabrication}
    \vspace{-0.25cm}
\end{figure}

The complete fabrication procedure is summarized in
Figure~\ref{fig:appendix_fabrication} and detailed below. All steps
are performed at room temperature. The entire
workflow, excluding print time, requires approximately 48 hours of
elapsed time, dominated by elastomer curing.

\paragraph{Step 1: Mold and spacer preparation
(Fig.~\ref{fig:appendix_fabrication}a).}
Three casting molds are generated from our parametric
CAD-to-mold pipeline (Section~\ref{sec:hardware}) and 3D-printed in
PLA. Mold~1 defines the bottom geometry of the elastomer pad and
incorporates precisely located protrusions that form blind
magnet-seating cavities during the first pour. Mold~2 and Mold~3 are two-part halves that together form the secondary
mold for the second pour; the split design allows the partially-cured
elastomer from Mold~1 to be inserted before the halves are assembled.
A PLA
spacer is separately designed around the magnetometer PCB footprint
and 3D-printed; it mechanically offsets the cured elastomer pad from
the surface-mount components.

\paragraph{Step 2: Elastomer preparation and first pour
(Fig.~\ref{fig:appendix_fabrication}b--c).}
XP-565 two-part silicone elastomer is mixed at a 10:1 base-to-activator
ratio by weight (6\,g base, 0.6\,g activator). The mixture is stirred
thoroughly and then degassed in a vacuum chamber (${\approx}$0.6\,bar below
ambient) for 20 minutes to eliminate entrained air bubbles. The
degassed mixture is poured into Mold~1 and allowed to partially cure
for 6.5 hours at room temperature, reaching approximately 40\% of the
full cure state (the rated full-cure time for XP-565 is 16 hours). At
this stage the elastomer is firm enough to hold shape but still
sufficiently tacky to bond to a subsequent pour.

\paragraph{Step 3: Magnet encapsulation and second pour
(Fig.~\ref{fig:appendix_fabrication}d--e).}
A fresh batch of XP-565 is prepared (4\,g base, 0.4\,g activator,
10:1 ratio) and degassed identically. The partially-cured elastomer is
demolded from Mold~1 and transferred to Mold~2. Mold~2 and Mold~3 are
two-part halves that together form the complete secondary mold cavity.
The split design is necessary to allow the partially-cured elastomer to
be inserted; a single-piece mold of this geometry would not permit
loading. With the elastomer seated in Mold~2, nine cylindrical
neodymium magnets ($\varnothing$\,2\,mm $\times$ 1\,mm height) are
pressed into the blind cavities formed during the first pour,
establishing the $3\!\times\!3$ array at their nominal positions. The
two halves are then assembled around the elastomer and the fresh
mixture is poured in until it reaches the fill line (indicated by the
red arrow in Fig.~\ref{fig:appendix_fabrication}e), permanently
encapsulating the magnets between the two elastomer layers. The
assembly is allowed to cure fully for 16 hours at room temperature.

\paragraph{Step 4: Adhesion treatment and bonding to spacer
(Fig.~\ref{fig:appendix_fabrication}f--g).}
Once fully cured, the elastomer pad is removed from the molds. Loctite
SF\,770 adhesion promoter is applied uniformly to the rear face of
the pad and allowed to flash off for one minute, priming the silicone
surface for cyanoacrylate bonding. The primed surface is then bonded
to the 3D-printed PLA spacer using Loctite 406 cyanoacrylate
adhesive. The bond is allowed to set for 1 minute before handling.

\paragraph{Step 5: Protective skin application and final assembly
(Fig.~\ref{fig:appendix_fabrication}h).}
A black anti-slip protective skin is laminated onto the active contact
surface of the elastomer pad to improve grip on smooth objects and
protect the elastomer from abrasive wear. The bonded assembly is left
undisturbed for 24 hours for complete adhesive curing. The finished
sensor pad is then fastened to the robot finger mount and the
magnetometer PCB using standard M2 machine screws, completing the
assembly.

To support reproducibility and community adoption, we release the
full parametric CAD-to-mold pipeline and the PCB board layout for
the $3{\times}3$ magnetometer array. Researchers can use the
pipeline to regenerate the casting molds for the exact configuration
used in our experiments, or modify the dimension and magnet-layout
parameters to customize the sensor geometry. The released PCB
layout can be submitted directly to a standard PCB fabrication
service for board ordering without requiring additional hardware design tooling.

\subsection{Sensor Characterization}
\label{app:hardware:characterization}

\begin{figure}[b]
    \centering
    \includegraphics[width=0.99\linewidth]{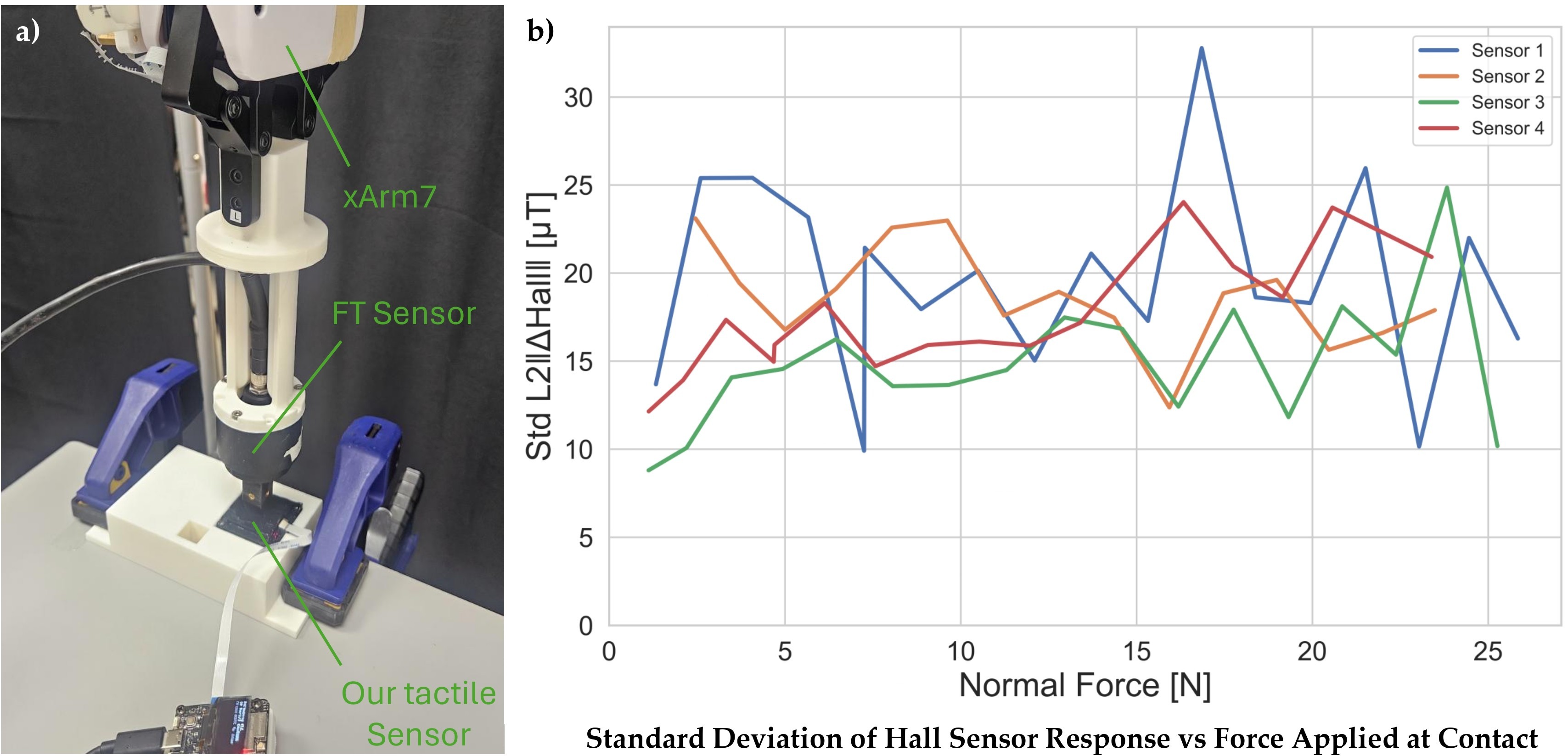}
    \caption{
    \footnotesize{
    \textbf{Sensor characterization experiments.}
    \textbf{(a) }Experiment setup: UFactory xArm7 equipped with BOTA Force-Torque sensor and our custom tactile sensor, used to collect ground-truth contact normal force data for characterizing the tactile sensor's response.
    \textbf{(b) }Standard deviation of the 4 independently fabricated sensor readings under different contact normal forces.}
    }
    \label{fig:appendix_tactile_study}
    \vspace{-0.25cm}
\end{figure}

We characterize the sensor along four axes: normal-force response and
consistency across independently fabricated units, directional shear
response, load--unload hysteresis, and long-duration baseline
stability. The normal-force study uses four sensors independently
fabricated following the procedure in
Appendix~\ref{app:hardware:fab}, each with the same $3{\times}3$
magnet-and-magnetometer configuration; all four units were fabricated
from independent 3D prints of the same parametric molds and different
elastomer batches by the same operator, with no post-fabrication
tuning or per-sensor calibration. The shear, hysteresis, and drift
studies use one representative unit (100\,Hz; 20\,Hz for the drift hold).

\paragraph{Experimental setup.}
All characterization experiments were conducted on a UFactory xArm7
manipulator equipped with a BOTA BFT-ROKA-SER-M8 six-axis
force-torque (FT) sensor mounted on the gripper fingers, providing
ground-truth contact force measurements
(Figure~\ref{fig:appendix_tactile_study}a). A flat $14\,\text{mm}
\times 14\,\text{mm}$ PLA probe was rigidly attached to the FT sensor
and brought into contact with the tactile sensor under position
control, following a study-specific motion protocol (stepped normal
indentation, tangential sweeps at fixed indentation, cyclic loading,
or a stationary hold), described with the corresponding results
below. In every experiment the raw 3-axis flux readings of all nine
magnetometers and the ground-truth FT wrench were recorded
simultaneously.

\paragraph{Response metrics.}
Throughout, $\mathbf{s}_i = \tilde{\mathbf{s}}_i - \mathbf{s}_i^0
\in \mathbb{R}^3$ denotes the offset-corrected flux of the $i$-th
magnetometer ($\tilde{\mathbf{s}}_i$ is the raw reading and
$\mathbf{s}_i^0$ is the per-trial resting offset recorded prior to
contact), in $\mu\text{T}$, and
$\mathbf{s} = [\mathbf{s}_1^\top, \ldots, \mathbf{s}_9^\top]^\top
\in \mathbb{R}^{27}$ their concatenation. The normal-force and
hysteresis studies report the aggregate response
$r = \|\mathbf{s}\|_2 \in \mathbb{R}$, which captures the total
contact-induced deflection summed across all nine magnetometers; the
shear study examines the in-plane components of individual
$\mathbf{s}_i$, and the drift study the per-channel deviations
directly.

\paragraph{Normal-force response and consistency across units.}
The probe is pressed perpendicularly onto the center of each of the
four units in $0.1$\,mm increments from zero contact to a maximum
indentation depth of $1.6$\,mm, with 20 repeat trials per depth.
Figure~\ref{fig:hardware_setup}e shows the mean response curve of
each of the four sensors individually as a function of applied normal
force, with per-sensor standard deviation shown as a transparent
band. The standard deviation is sufficiently small that the bands
are not visible at the full plot scale; an inset provides a zoomed
view to make the within-sensor variation legible.
Figure~\ref{fig:appendix_tactile_study}b shows the per-sensor
trial-to-trial standard deviation of $r$ across the 20 repeat
measurements at each normal force level.

Two properties are evident. First, all four sensors exhibit a
monotonically increasing and smooth response across the full
indentation range, confirming that raw magnetic flux provides a
consistent and unambiguous contact signal without force calibration.
Second, the trial-to-trial standard deviation within each sensor
remains low throughout, in the $10$--$30\,\mu\text{T}$ range,
indicating stable repeatability regardless of applied force magnitude.

The four mean response curves are not identical: unit-to-unit
differences in absolute response reach approximately 20\%, reflecting
minor geometric variation from the manual fabrication process (e.g.,
magnet seating depth, elastomer thickness). Importantly, this
inter-sensor offset does not affect our pipeline: each sensor unit
is paired with demonstrations collected by that same unit, so
per-channel normalization using dataset statistics (described in
Appendix~\ref{app:realworld:setup}) absorbs the constant scale and
bias difference between units without requiring any dedicated
calibration hardware.

\begin{figure}[t]
    \centering
    \includegraphics[height=5.2cm]{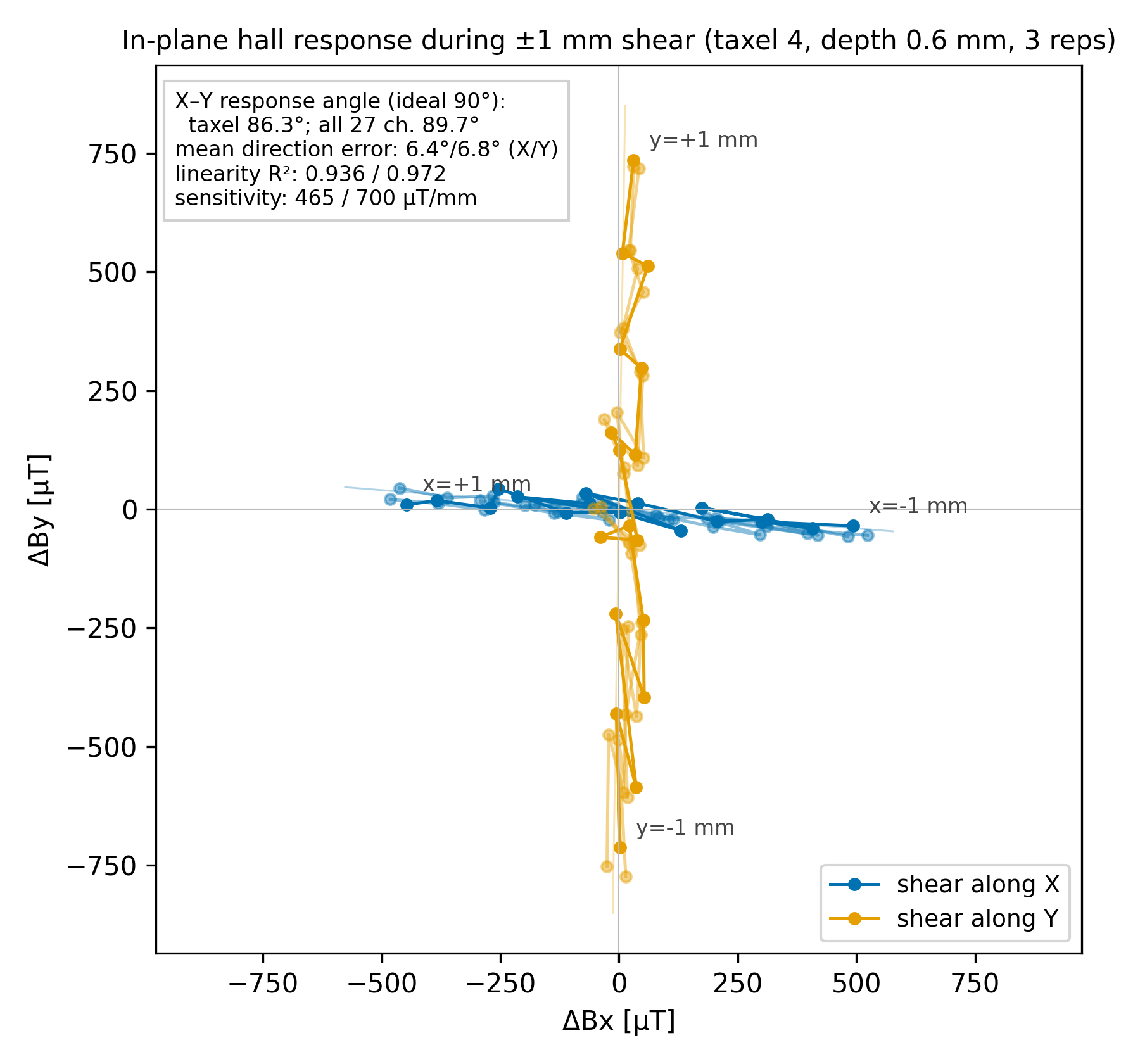}\hfill
    \includegraphics[height=5.2cm]{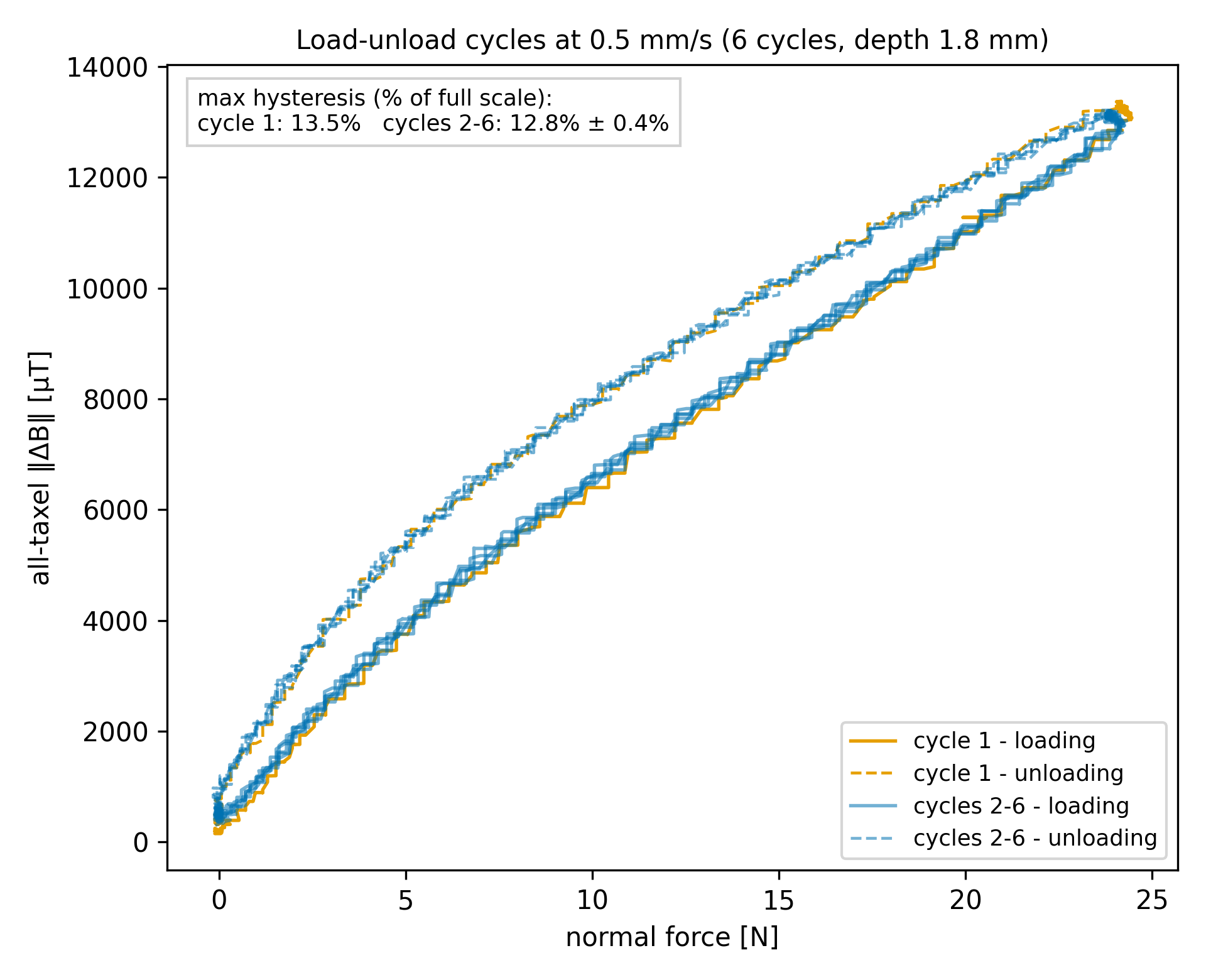}
    \caption{
    \footnotesize{
    \textbf{Directional shear response and hysteresis.}
    \textbf{Left:} in-plane flux change of the center taxel during
    $\pm1$\,mm tangential sweeps along X and Y at $0.6$\,mm
    indentation (3 repetitions): orthogonal, monotonic,
    sign-symmetric response trajectories.
    \textbf{Right:} aggregate response $r$ versus normal force over
    six consecutive load--unload cycles at $0.5$\,mm/s; maximum
    loading--unloading separation is $\le$14\% of full scale and
    identical across cycles.}
    }
    \label{fig:appendix_directional}
\end{figure}

\paragraph{Shear response and directional accuracy.}
To verify that the measurements carry directional information, the
probe is pressed to a fixed indentation of $0.6$\,mm
($\approx$6.5\,N) at the sensor center and translated tangentially to
$\pm 1$\,mm along the $\pm$X and $\pm$Y axes in $0.25$\,mm steps
($2$\,mm/s, $1.5$\,s settled hold per step, 3 repetitions per axis).
Figure~\ref{fig:appendix_directional} (left) plots the in-plane flux
change of the center taxel: the X- and Y-shear sweeps trace
monotonic, sign-symmetric, nearly orthogonal trajectories. The
response is $465$ / $700\,\mu$T/mm along X/Y with $R^2 = 0.936$ /
$0.972$ (linear fit pooled over all repetitions and both sweep
directions), cross-axis sensitivity is $\le$10\%, and a single
settled reading indicates the shear direction to within $7^\circ$ on
average. The X- and Y-shear response directions are $86.3^\circ$
apart at the center taxel and $89.7^\circ$ apart when measured
between the 27-channel response vectors (ideal: $90^\circ$),
confirming that orthogonal shears produce nearly orthogonal
array-level signatures. The cross-axis figure is an upper bound: it
includes leakage from a small ($\approx$3$^\circ$) rotation between
the sensor and motion axes, estimated from the common offset of both
fitted directions.

\paragraph{Hysteresis.}
Six consecutive load--unload cycles to $24$\,N ($0.5$\,mm/s,
$1.8$\,mm depth) are applied at the sensor center
(Figure~\ref{fig:appendix_directional}, right). The aggregate
response $r$ forms a closed loop with maximum loading--unloading
separation of $12.4$--$13.6\%$ of full scale per cycle ($12.9\%$
mean), measured at matched force over $10$--$90\%$ of the force
range; this figure includes the rate-dependent viscoelastic lag at
the applied speed. The loop is unchanged from the first cycle onward,
indicating no conditioning effect or cumulative drift under repeated
loading.

\begin{figure}[t]
    \centering
    \includegraphics[width=0.99\linewidth]{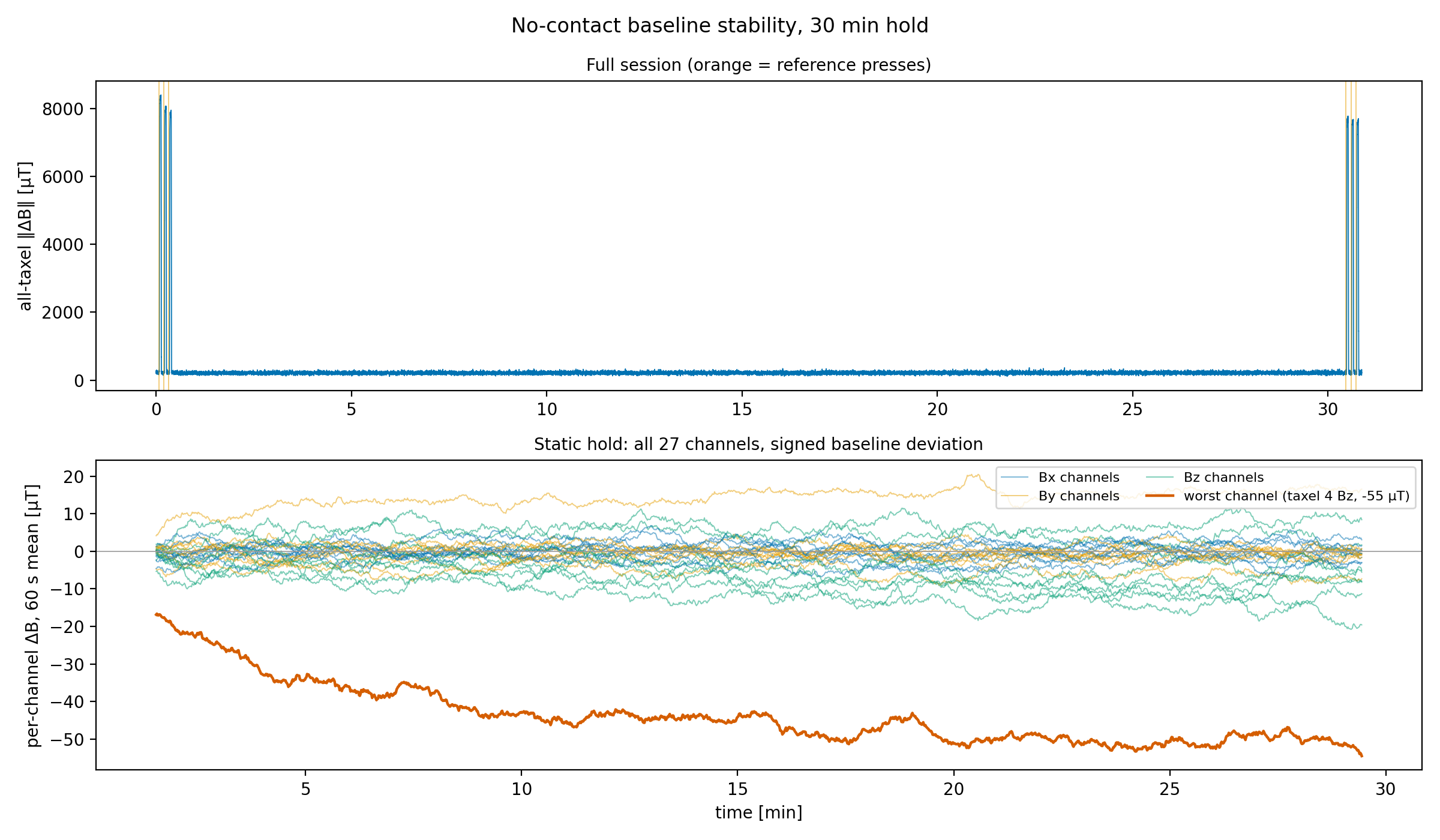}
    \caption{
    \footnotesize{
    \textbf{Baseline stability over a 30-minute stationary hold.}
    \textbf{Top:} aggregate response over the full session; orange
    lines mark the reference presses bracketing the hold.
    \textbf{Bottom:} signed per-channel baseline deviation (60\,s
    means) of all 27 channels during the hold; 26 of 27 channels stay
    within $\pm20\,\mu$T, and the largest excursion ($-55\,\mu$T,
    center-taxel normal axis) is a settling transient of $0.7\%$ of a
    typical press signal.}
    }
    \label{fig:appendix_drift}
\end{figure}

\paragraph{Baseline drift.}
The sensor is left untouched for 30 minutes, bracketed by three
reference presses before and after
(Figure~\ref{fig:appendix_drift}). Relative to the first minute, 26
of 27 channels remain within $\pm 20\,\mu$T (60\,s means) for the
entire hold; the largest excursion, on the center taxel's normal
axis, is a $-55\,\mu$T settling transient that converges within
$\sim$10 minutes, consistent with elastomer relaxation after the
preceding presses. This worst-case excursion is $0.7\%$ of a typical
$8{,}100\,\mu$T press signal, below the per-sample noise, and the
reference press amplitudes before and after the hold agree within
$5.2\%$. In deployment, the per-trial resting offset
$\mathbf{s}_i^0$ removes such slow baseline shifts by construction. (In this
recording, $0.04\%$ of frames contained isolated single-channel
serial read errors and were replaced by the channel's rolling
median.)

\newpage

\section{Details of BC Transformer on LIBERO}
\label{app:bct}

\subsection{BC-Transformer Details}
\label{app:impl:bct}

For results described in \cref{sec:exp:libero}, we train with the default BC-Transformer policy implementation in LIBERO~\cite{liu2023libero}. Hyperparameters are listed in \cref{tab:bct-hparams}.

\begin{table}[h]
\centering
\small
\setlength{\tabcolsep}{6pt}
\renewcommand{\arraystretch}{1.15}
\caption{BC-Transformer training configuration for LIBERO multitask experiments.}
\label{tab:bct-hparams}
\begin{tabular}{ll}
\toprule
\textbf{Component / Hyperparameter} & \textbf{Value} \\
\midrule
\multicolumn{2}{l}{\textit{Architecture}} \\
Visual backbone                 & ResNet-18 (random init, \texttt{remove\_layer\_num=4}, no stride change) \\
Language fusion                 & FiLM conditioning on visual features \\
Language encoder                & 1-layer MLP (input $768$, hidden $128$, output $128$) \\
Transformer layers              & $4$ \\
Hidden / embed dimension        & $64$ (token embed); MLP hidden $256$ \\
Attention heads                 & $6$ (per-head dim $64$) \\
Action head                     & GMM head, $2$ MLP layers, hidden $1024$, $5$ modes, softplus, \texttt{min\_std}$=10^{-4}$ \\
Proprio inputs                  & gripper\_states, joint\_states (\texttt{eye\_in\_hand} + \texttt{agentview} RGB) \\
\midrule
\multicolumn{2}{l}{\textit{Optimization}} \\
Optimizer                       & AdamW ($\beta_1{=}0.9$, $\beta_2{=}0.999$) \\
Learning rate                   & $1\!\times\!10^{-4}$ (cosine annealing to $\eta_{\min}{=}1\!\times\!10^{-5}$) \\
Weight decay                    & $1\!\times\!10^{-4}$ \\
Dropout                         & $0.1$ (transformer) \\
Gradient clipping               & $100.0$ \\
Batch size                      & $32$ \\
Training epochs                 & $20$ \\
\midrule
\multicolumn{2}{l}{\textit{Sequence \& action}} \\
Sequence length (context)       & $10$ frames (\texttt{transformer\_max\_seq\_len}$=10$, \texttt{seq\_len}$=10$) \\
Frame stack                     & $1$ \\
Action chunking                 & None (single-step action prediction) \\
Action scale                    & $1.0$ \\
Image resolution                & $128\!\times\!128$ \\
Data augmentation               & batch-wise color jitter + translation aug \\
\midrule
\multicolumn{2}{l}{\textit{Lifelong configuration}} \\
\texttt{lifelong.algo}          & \texttt{Multitask} (all tasks jointly via \texttt{learn\_all\_tasks}) \\
\texttt{eval\_in\_train}        & \texttt{false} \\
\bottomrule
\end{tabular}
\end{table}

\subsection{LIBERO Main Results}
\label{app:main_libero}


\begin{table}[b]
\centering
\small
\caption{BC-Transformer success rates (\%) on LIBERO suites across
camera configurations ($n{=}3$ seeds).}
\label{tab:libero_main}
\vspace{0.1cm}
\begin{tabularx}{\linewidth}{l *{5}{Y}}
\toprule
& \multicolumn{2}{c}{1-view (agentview)} & \multicolumn{3}{c}{2-view (agentview + wrist)} \\
\cmidrule(lr){2-3} \cmidrule(lr){4-6}
Suite & Plain & \ACRO & Plain & Proprio & \ACRO \\
\midrule
OBJECT  & $90.7 \pm 4.4$ & $\mathbf{95.0 \pm 0.9}$ & $82.8 \pm 7.0$ & $80.0 \pm 2.0$ & $\mathbf{98.2 \pm 1.2}$ \\
GOAL    & $79.3 \pm 3.2$ & $\mathbf{91.7 \pm 0.3}$ & $86.8 \pm 4.1$ & $81.5 \pm 4.3$ & $\mathbf{92.8 \pm 2.5}$ \\
SPATIAL & $79.0 \pm 2.0$ & $\mathbf{90.5 \pm 0.4}$ & $84.8 \pm 4.0$ & $80.0 \pm 1.4$ & $\mathbf{92.3 \pm 2.7}$ \\
LONG    & $41.8 \pm 6.1$ & $\mathbf{63.8 \pm 1.2}$ & $52.8 \pm 2.1$ & $42.8 \pm 5.1$ & $\mathbf{86.5 \pm 2.3}$ \\
\midrule
Average & $72.7$         & $\mathbf{85.2}$         & $76.8$         & $71.1$         & $\mathbf{92.5}$ \\
\bottomrule
\end{tabularx}
\end{table}

Table~\ref{tab:libero_main} reports BC-Transformer success rates
across all four LIBERO suites and two camera configurations
($n{=}3$ seeds), aligned with the bar plots in \cref{fig:bct-main}. We discuss a few observations beyond the Q1/Q2 summary in
the main paper.

\paragraph{The overlay benefit grows with task difficulty.}
On the easier OBJECT, GOAL, and SPATIAL suites, the 2-view \ACRO\
gain over plain images is between $+6$pp and $+15$pp. On the
harder LIBERO\_LONG suite, the gain widens to $+33.7$pp. The
pattern is consistent with our per-task analysis
(Appendix~\ref{app:per_task_libero}): overlay benefit correlates
strongly with task headroom ($r = 0.924$), so suites where
vision-only performance is lowest see the largest improvement.

\paragraph{Overlays stabilize as well as improve.}
Across the four suites, \ACRO\ produces tighter cross-seed
standard deviations than the plain or Proprio baselines (median
SD $2.4\%$ for \ACRO\ 2v vs. $4.1\%$ for plain 2v and $3.2\%$ for
Proprio). On the OBJECT 2v plain condition the standard deviation
reaches $7.0\%$, driven by a single high-variance task analyzed in
Section~\ref{app:per_task_libero}; \ACRO\ 2v on the same suite
has SD $1.2\%$. Contact overlays appear to both raise mean
performance and reduce seed-to-seed variability.

\paragraph{Proprio underperforms plain on every suite.}
The Proprio baseline provides an aggregated 12-D contact wrench
concatenated to the state vector. It underperforms the plain 2-view
baseline on every suite ($-2.8$pp on OBJECT, $-5.3$pp on GOAL,
$-4.8$pp on SPATIAL, $-10.0$pp on LONG), and underperforms even
the 1-view plain baseline ($71.1\%$ avg vs.\ $72.7\%$). Adding this
compact state-vector representation of contact consistently hurts
BC-Transformer performance in this setup. By contrast, image-space
rendering yields a $+15.7$pp average improvement over plain 2-view
and a $+21.4$pp gap over Proprio. These results show that, for
BC-Transformer, image-space rendering is substantially more effective
than this compact state-vector concatenation. This comparison does
not establish superiority over learned tactile encoders or other
feature-level fusion mechanisms.

\paragraph{The wrist camera helps \ACRO\ universally but hurts
plain on OBJECT.}
Adding the wrist camera to the plain baseline yields modest
average gains ($72.7 \to 76.8$, $+4.1$pp) but regresses
substantially on OBJECT ($90.7 \to 82.8$, $-7.9$pp; see
Section~\ref{app:per_task_libero} for the task-level analysis).
\ACRO\ benefits from the wrist camera on every suite without
exception ($+7.3$pp average), including OBJECT ($+3.2$pp). The
wrist camera and the overlay are complementary rather than
substitutable: the overlay rescues the OBJECT 2-view condition
from the training-instability regime affecting the plain
counterpart.

\subsection{Per Task Analysis on LIBERO}
\label{app:per_task_libero}
This appendix
characterizes the per-task structure of overlay gains across all 40
LIBERO tasks.

\paragraph{Rescue Outliers}
On each suite, a small number of tasks show particularly large overlay
gains. These outliers all involve unusual or precarious object
configurations where visual perception alone is unreliable and
per-finger contact arrows fill in the missing information.
Table~\ref{tab:rescue_outliers} reports the most prominent example from
each suite.

\begin{table}[h]
\centering
\caption{Rescue outliers: tasks where Avg.\ Arrow produces the
largest gain over plain 2-view within each suite. All four outliers
involve grasp configurations that are visually ambiguous (slippery,
small, occluded, or perched objects), where the contact arrow provides
information not available from pixels alone.}
\label{tab:rescue_outliers}
\setlength{\tabcolsep}{6pt}
\begin{tabular}{llccc}
\toprule
Suite & Task & Plain 2v & \ACRO\ 2v & $\Delta (pp)$ \\
\midrule
OBJECT  & butter $\to$ basket            & $48\%$  & $98\%$  & $+50$ \\
GOAL    & wine $\to$ top of cabinet      & $83\%$  & $100\%$ & $+17$ \\
SPATIAL & bowl on ramekin $\to$ plate    & $62\%$  & $93\%$  & $+32$ \\
LONG    & cheese + butter $\to$ basket   & $8\%$   & $95\%$  & $+87$ \\
\bottomrule
\end{tabular}
\end{table}

\paragraph{OBJECT 2-View Training Instability}
Table~\ref{tab:libero_main} reports an anomalous 2-view plain result
on \texttt{LIBERO\_OBJECT} ($82.8 \pm 7.0\%$), which underperforms
the 1-view counterpart ($90.7 \pm 4.4\%$) despite positive
wrist-camera gains on all three other suites. The regression is
driven largely by task 6 (\texttt{butter $\to$ basket}), whose
success rates across three seeds
show substantial seed-to-seed variability indicative of training instability;
excluding task 6, the 2-view plain average on the remaining 9 tasks
is $87\%$, consistent with 1-view. We attribute this to
under-training at 20 epochs under the \texttt{lifelong=multitask}
protocol on a visually low-contrast scene, but report the result
honestly to preserve protocol consistency across cells. The
regression does not affect our central claim: \ACRO\ on OBJECT
($98.2 \pm 1.2\%$) exceeds both plain baselines by margins well
outside seed variance, and Proprio ($80.0\%$) underperforms even
the 1-view plain baseline.

\subsection{Task-Level Correlation Analysis}
\label{app:per_task_libero:correlations}

\paragraph{Task properties.}
For each of the 40 LIBERO tasks, we compute the following properties
from the demonstration data and the contact extraction:
\begin{itemize}[leftmargin=1em,itemsep=0.1em,topsep=0.2em]
    \item \textbf{Headroom}: $1 - \text{plain 2v success rate}$,
    capturing how much room the task leaves for improvement.
    \item \textbf{\#grasps}: median number of gripper-close events per
    demonstration trajectory (across 50 demos per task).
    \item \textbf{Trajectory length}: mean trajectory length in
    environment steps, across 50 demos per task.
    \item \textbf{\%contact}: mean fraction of trajectory steps with
    non-zero gripper-finger contact wrench.
    \item \textbf{\#contact onsets}: mean number of contact-state
    transitions (zero $\to$ non-zero wrench) per trajectory.
\end{itemize}

\paragraph{Outcome variable.}
$\Delta = \text{\ACRO\ 2v} - \text{plain 2v}$ per task, computed as
the difference of seed-averaged success rates.

\paragraph{Raw and partial correlations.}
We report Pearson correlations between each task property and $\Delta$
across the 40 tasks. To control for the trivial dependence on headroom
(low-performing tasks have more room to improve), we additionally
report partial Pearson correlations after regressing out headroom from
both $\Delta$ and the predictor. Full results are in
Table~\ref{tab:correlations}.

\begin{table}[h]
\centering
\caption{Per-task correlations between $\Delta = $ \ACRO\ 2v $-$ plain
2v and task properties, pooled across all 40 LIBERO tasks. Raw Pearson
correlations are dominated by headroom (overlays help most where vision
fails). After controlling for headroom via partial correlation, only
\textit{\#contact onsets} and \textit{trajectory length} retain
independent predictive signal.}
\label{tab:correlations}
\setlength{\tabcolsep}{6pt}
\begin{tabular}{lcc|cc}
\toprule
& \multicolumn{2}{c|}{Raw Pearson} & \multicolumn{2}{c}{Partial (controlling for headroom)} \\
Feature & $r$ & $p$ & $r$ & $p$ \\
\midrule
Headroom            & $+0.924$ & $<10^{-3}$ & --- & --- \\
\#grasps            & $+0.633$ & $<10^{-3}$ & $-0.094$ & $0.562$ \\
Trajectory length   & $+0.458$ & $0.003$    & $-0.368$ & $0.019$ \\
\%contact           & $+0.360$ & $0.022$    & $+0.214$ & $0.186$ \\
\#contact onsets    & $+0.329$ & $0.038$    & $-0.571$ & $<10^{-3}$ \\
\bottomrule
\end{tabular}
\end{table}

\subsection{Additional Comparisons}
\label{app:additional}

\subsubsection{Effect of Camera Configuration on Plain Baseline}
\label{app:additional:camera}

Table~\ref{tab:libero_main} reports both 1-view and 2-view plain
baselines, but the per-suite comparison is easier to see in
isolation. Table~\ref{tab:camera_config_plain} reports the
per-suite 1-view $\to$ 2-view change for the plain (no-overlay)
baseline. Three of four suites show the expected positive gain
from adding a wrist camera (GOAL $+7.5$pp, SPATIAL $+5.8$pp,
LONG $+11.0$pp), while OBJECT regresses $-7.9$pp. The OBJECT
regression is driven by a single high-variance task.

\begin{table}[h]
\centering
\caption{Effect of camera configuration on the plain (no-overlay)
BC-Transformer baseline. Mean success rate (\%) over 3 seeds per
cell, 10 tasks per suite. Three of four suites benefit from the
wrist camera; OBJECT regresses.}
\label{tab:camera_config_plain}
\setlength{\tabcolsep}{8pt}
\begin{tabular}{lccc}
\toprule
Suite     & 1-view Plain & 2-view Plain & $\Delta$ (2v $-$ 1v) \\
\midrule
OBJECT    & $90.7$       & $82.8$       & $\mathbf{-7.9}$ \\
GOAL      & $79.3$       & $86.8$       & $+7.5$ \\
SPATIAL   & $79.0$       & $84.8$       & $+5.8$ \\
LONG      & $41.8$       & $52.8$       & $+11.0$ \\
\midrule
Average   & $72.7$       & $76.8$       & $+4.1$ \\
\bottomrule
\end{tabular}
\end{table}

In contrast, \ACRO\ benefits from the wrist camera on every
suite, including OBJECT (\cref{tab:libero_main}): the wrist
camera and the overlay are complementary rather than substitutable,
and the overlay rescues the OBJECT 2-view condition from the
training-instability regime affecting the plain 2-view baseline
(98.2\% with \ACRO\ vs.\ 82.8\% plain 2v).

\subsubsection{Seed-Level Results}
\label{app:additional:seeds}

Table~\ref{tab:libero_main} reports mean $\pm$ standard deviation
over 3 seeds. Tables~\ref{tab:seedlevel_plain},
\ref{tab:seedlevel_proprio}, and \ref{tab:seedlevel_acro} report
per-seed success rates for the three headline 2-view conditions
(Plain, Proprio, \ACRO).

\begin{table}[h]
\centering
\caption{Per-seed success rates (\%) for the \textbf{Plain 2-view}
BC-Transformer baseline.}
\label{tab:seedlevel_plain}
\setlength{\tabcolsep}{8pt}
\begin{tabular}{lccccc}
\toprule
Suite     & Seed 12345 & Seed 23456 & Seed 34567 & Mean & SD \\
\midrule
OBJECT    & $84.5$ & $73.5$ & $90.5$ & $82.8$ & $7.0$ \\
GOAL      & $90.0$ & $89.5$ & $81.0$ & $86.8$ & $4.1$ \\
SPATIAL   & $90.5$ & $82.0$ & $82.0$ & $84.8$ & $4.0$ \\
LONG      & $50.5$ & $52.5$ & $55.5$ & $52.8$ & $2.1$ \\
\bottomrule
\end{tabular}
\end{table}

\begin{table}[h]
\centering
\caption{Per-seed success rates (\%) for the \textbf{Proprio 2-view}
baseline (Plain RGB + 12-D contact wrench concatenated to the
proprioceptive state).}
\label{tab:seedlevel_proprio}
\setlength{\tabcolsep}{8pt}
\begin{tabular}{lccccc}
\toprule
Suite     & Seed 12345 & Seed 23456 & Seed 34567 & Mean & SD \\
\midrule
OBJECT    & $82.5$ & $77.5$ & $80.0$ & $80.0$ & $2.0$ \\
GOAL      & $84.0$ & $75.5$ & $85.0$ & $81.5$ & $4.3$ \\
SPATIAL   & $82.0$ & $79.0$ & $79.0$ & $80.0$ & $1.4$ \\
LONG      & $50.0$ & $38.5$ & $40.0$ & $42.8$ & $5.1$ \\
\bottomrule
\end{tabular}
\end{table}

\begin{table}[h]
\centering
\caption{Per-seed success rates (\%) for \textbf{\ACRO\ 2-view}
(default Avg.\ Arrow overlay).}
\label{tab:seedlevel_acro}
\setlength{\tabcolsep}{8pt}
\begin{tabular}{lccccc}
\toprule
Suite     & Seed 12345 & Seed 23456 & Seed 34567 & Mean & SD \\
\midrule
OBJECT    & $99.0$ & $99.0$ & $96.5$ & $98.2$ & $1.2$ \\
GOAL      & $96.0$ & $92.5$ & $90.0$ & $92.8$ & $2.5$ \\
SPATIAL   & $96.0$ & $91.5$ & $89.5$ & $92.3$ & $2.7$ \\
LONG      & $84.0$ & $86.0$ & $89.5$ & $86.5$ & $2.3$ \\
\bottomrule
\end{tabular}
\end{table}

\paragraph{Seed-to-seed variability.}
Plain 2v shows the largest per-seed variation, with the OBJECT
suite spanning a 17.0\,pp range across seeds (73.5\% to 90.5\%)
and producing a standard deviation of $7.0\%$. 
\ACRO\ 2v has the
narrowest cross-seed spreads on every suite (SD between $1.2\%$
and $2.7\%$), indicating that the overlay both improves average
performance and stabilizes training across random seeds. Proprio
2v sits between the two, with relatively higher variance on the
LONG suite ($\text{SD}=5.1\%$).

\newpage

\section{miniVLA on LIBERO Details}
\label{app:minivla_libero}

\subsection{miniVLA Pretraining on LIBERO\_90 (Contact-Overlay)}
\label{app:minivla_pretraining}

\paragraph{Backbone.}
\texttt{prism-qwen25-extra-dinosiglip-224px+0\_5b}: Qwen2.5-0.5B
LLM with a fused DINOv2 + SigLIP $224{\times}224$ vision encoder.
The full VLM is unfrozen (no frozen LLM or vision backbone, no
LoRA). Actions are emitted through the Qwen \emph{extra} action
tokenizer (256 extra vocab tokens, leaving the language vocab
intact).



\paragraph{Optimization.}
\begin{itemize}[leftmargin=1em,itemsep=0.1em,topsep=0.2em]
  \item \textbf{Train strategy:} DDP
        (\texttt{train\_strategy=ddp}, custom registration on top
        of openvla-mini's FSDP-only default; grad-checkpointing path
        and submodule unwrap patched locally).
  \item \textbf{Batch:} 4 GPUs $\times$ per-device batch 8 = global
        batch 32.
  \item \textbf{Learning rate:} $2{\times}10^{-5}$,
        \texttt{linear-warmup + cosine-decay},
        \texttt{warmup\_ratio=0.0},
        \texttt{weight\_decay=0.0}, \texttt{max\_grad\_norm=1.0}.
  \item \textbf{Precision:} mixed precision with
        \texttt{reduce\_in\_full\_precision=true}; gradient
        checkpointing enabled.
  \item \textbf{Data loader:} shuffle buffer 256{,}000; seed 7.
  \item \textbf{Budget:} \texttt{max\_steps=200{,}000}
        ($\approx$11 epochs on the 90-suite mix); checkpoint every
        25{,}000 steps.
\end{itemize}

\begin{figure}[hbt]
    \centering
    \includegraphics[width=\linewidth]{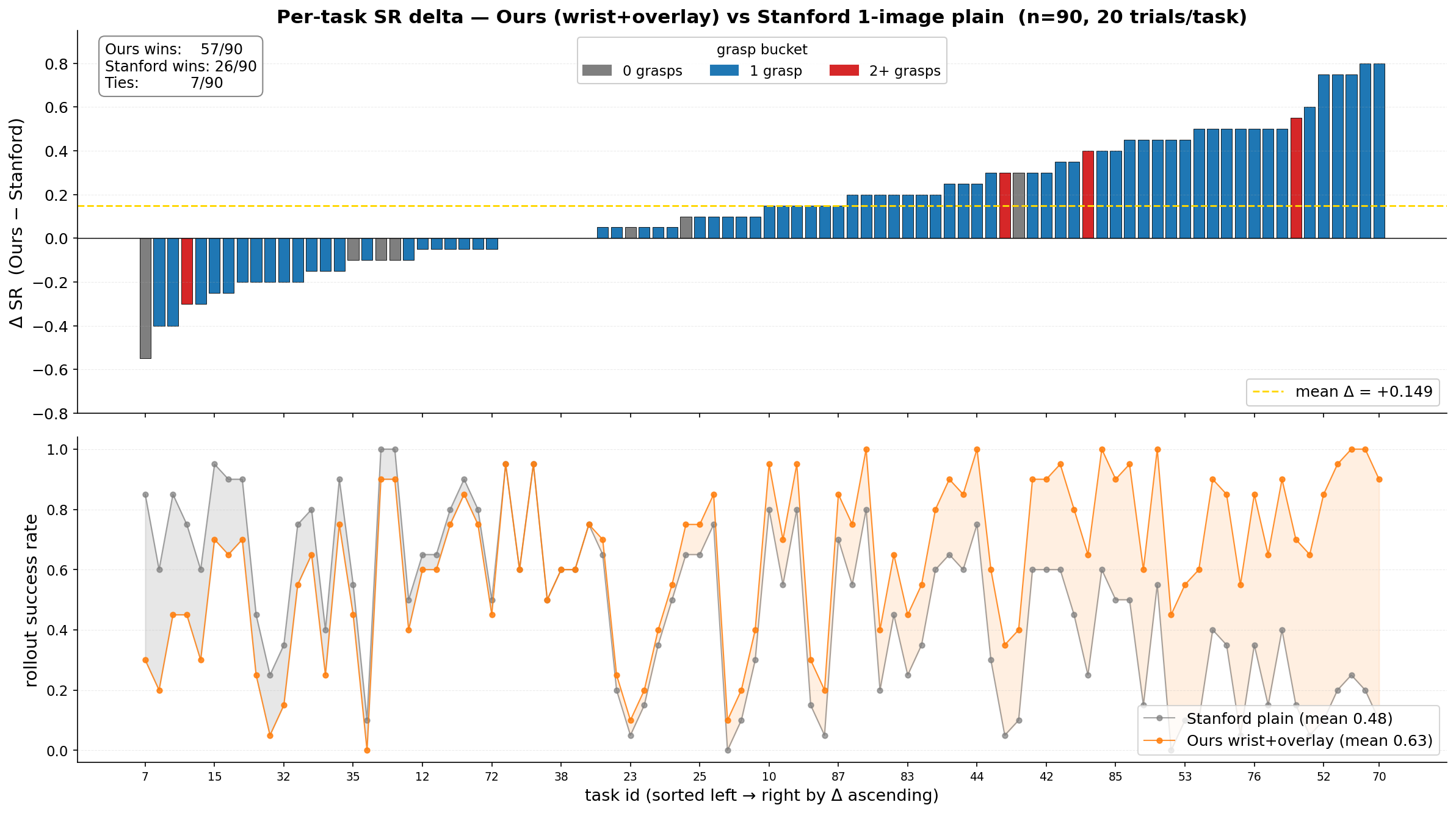}
    \caption{Per-task success rates on LIBERO\_90: \ACRO-miniVLA
    (contact-overlay pretraining) vs.\ the re-evaluated Stanford
    miniVLA~\cite{belkhale2024minivla} baseline.}
    \label{fig:libero90_pertask}
\end{figure}

\paragraph{Comparison with the released Stanford miniVLA.}
We trained \ACRO-miniVLA on 4$\times$ NVIDIA L40S GPUs at  batch 8
(global batch 32) for 200k steps ($\approx$11 epochs), reaching a
final training loss of $0.364$. The released Stanford
baseline~\cite{belkhale2024minivla} was trained on 8$\times$ NVIDIA A100
GPUs at per-device batch 32 (global batch 256) for 60k steps
($\approx$26 epochs) with a constant learning rate, reaching a
final loss of $0.129$. Despite the lower compute budget and weaker
training-loss convergence, head-to-head re-evaluation on all 90
LIBERO\_90 tasks shows \ACRO-miniVLA at $63.2\%$ success versus
the Stanford baseline at $48.3\%$ ($+14.9$pp; 20 trials per task).
\cref{fig:libero90_pertask} shows the per-task comparison.


\paragraph{Additional comparison.}
We additionally evaluate \ACRO-miniVLA under the VQ-h8 +
history=2 architecture used for the headline LIBERO\_90 result in
the original miniVLA work~\cite{belkhale2024minivla}. We fine-tune
the publicly released miniVLA checkpoint on contact-overlay
LIBERO\_90 data (batch size 8, two NVIDIA L40S GPUs). The
resulting model reaches $86.3\%$ mean success on the 90-task
suite (10 trials per task), while the released checkpoint obtains
$72.2\%$ under our identical evaluation pipeline, a $+14.1$pp
gain from contact overlays under matched conditions; for
reference, the original work reports $82\%$ on its own harness.
This demonstrates that contact overlays remain effective under a
different miniVLA architectural configuration.

\color{black}

\subsection{miniVLA Fine-Tuning Details}
\label{app:minivla_finetuning}

For the main-paper fine-tuning results
(Table~\ref{tab:minivla_finetuning}), we report mean \ACRO\
performance over five training seeds using a suite-specific training
recipe, while baseline results are single runs. For transparency,
this appendix additionally reports per-recipe comparisons to
characterize whether the observed gains depend on the fine-tuning
recipe.

\subsubsection{Recipes}
\label{app:minivla_recipes:recipes}

We fine-tune History-2 miniVLA~\cite{belkhale2024minivla} on
LIBERO\_90 with and without contact overlays. We evaluate two
principal recipes.

\paragraph{b8 1-epoch.}
A lightweight recipe used for early exploration: batch size 8,
single-GPU training, 1 epoch over the LIBERO\_90 demonstrations.
This recipe is fast to train and serves as a sanity check for
whether contact overlays help under a constrained training budget.

\paragraph{DDP-4 b16 cosine.}
A stronger distributed recipe used for the headline \ACRO\ results on
OBJECT, GOAL, and LONG: distributed data-parallel training across
4 GPUs, per-GPU batch size 16 (effective batch size 64), with a cosine
learning-rate schedule and weight decay. This recipe trains for
substantially longer wall-clock time than b8 1-epoch.

\paragraph{2-phase (SPATIAL only).}
For SPATIAL, we additionally evaluate a two-phase training procedure
for both the plain baseline and \ACRO. This recipe provides the
suite-specific SPATIAL results reported in
Table~\ref{tab:minivla_finetuning}: the plain baseline reaches
$71.5\%$, while \ACRO\ reaches $87.0 \pm 1.8\%$ over five training
seeds. For comparison, the recipe-diagnostic table below also reports
the single-run DDP-4 b16 cosine results for SPATIAL.

\subsubsection{Evaluation Protocol}
\label{app:minivla_recipes:eval}

Overlay-conditioned models are evaluated with contact overlays rendered
at resolution 256 and ROT180 preprocessing on both camera views to align
with the LIBERO\_90 convention. For the suite-specific \ACRO\
configurations reported in the main paper, we train five independent
seeds; baseline and alternative-recipe results are single runs.
Other evaluation parameters, including rollouts per task and the
success criterion, follow the protocol in Section~\ref{sec:exp:libero}.

\subsubsection{Per-Recipe Results}
\label{app:minivla_recipes:results}

Table~\ref{tab:minivla_recipe_diagnostic} reports per-recipe results
across all four LIBERO suites.

\begin{table}[h]
\small
\centering
\caption{Per-recipe comparison for History-2 miniVLA on LIBERO suites.
\textbf{Baseline}: clean non-contact fine-tuning with clean evaluation.
\textbf{\ACRO}: contact-overlay fine-tuning with overlay-augmented evaluation.
For the suite-specific recipes used in the main results, \ACRO performance
is reported as mean $\pm$ standard error over 5 training seeds; all other
entries are single runs. Baseline results are single runs.}
\label{tab:minivla_recipe_diagnostic}
\renewcommand{\arraystretch}{1.15}
\begin{tabularx}{\linewidth}{l l *{3}{Y}}
\toprule
\textbf{Suite} & \textbf{Recipe} & \textbf{Baseline} & \textbf{\ACRO} & \textbf{$\Delta$} \\
\midrule
OBJECT
& b8 1-epoch
& $47.0\%$
& $57.5\%$
& $+10.5$ pp \\

OBJECT
& DDP-4 b16 cosine
& $51.0\%$
& $70.6 \pm 4.6\%$
& $+19.6$ pp \\
\midrule

GOAL
& b8 1-epoch
& $48.5\%$
& $74.0\%$
& $+25.5$ pp \\

GOAL
& DDP-4 b16 cosine
& $37.4\%$
& $80.5 \pm 4.3\%$
& $+43.1$ pp \\
\midrule

SPATIAL
& 2-phase
& $71.5\%$
& $87.0 \pm 1.8\%$
& $+15.5$ pp \\

SPATIAL
& DDP-4 b16 cosine
& $55.0\%$
& $87.5\%$
& $+32.5$ pp \\
\midrule

LONG
& b8 1-epoch
& $32.5\%$
& $58.0\%$
& $+25.5$ pp \\

LONG
& DDP-4 b16 cosine
& $26.0\%$
& $66.6 \pm 2.2\%$
& $+40.6$ pp \\
\bottomrule
\end{tabularx}
\end{table}

\paragraph{Tactile gains are robust across recipes.}
Contact-overlay fine-tuning improves over non-contact fine-tuning
for every suite and training recipe evaluated. In the main table,
the five-seed \ACRO\ means improve over the strongest available
single-run baselines by $+19.6$ pp on OBJECT, $+32.0$ pp on GOAL,
$+15.5$ pp on SPATIAL, and $+34.1$ pp on LONG. The alternative
single-run recipe comparisons show the same trend, with gains
ranging from $+10.5$ pp to $+32.5$ pp.

\paragraph{Recipe-matched comparisons support the same trend.}
For OBJECT, the main-table comparison is already recipe-matched under
DDP-4 b16 cosine, yielding a $+19.6$ pp gain. For GOAL and LONG, the
same five-seed \ACRO\ means reported in the main table are $+43.1$ pp
and $+40.6$ pp above the corresponding DDP-4 single-run baselines,
respectively. The main table instead compares against the stronger
single-run baselines available for these suites, yielding the more
conservative gains of $+32.0$ pp and $+34.1$ pp. On SPATIAL, the
2-phase comparison used in the main table yields a $+15.5$ pp gain,
while the alternative single-run DDP-4 b16 cosine comparison yields
$+32.5$ pp. Together, these comparisons indicate that the improvement
from contact overlays is not specific to a particular fine-tuning
recipe.



\subsection{Image Orientation Probe}
\label{app:orientation}

For the miniVLA pretraining comparison reported in
Section~\ref{sec:exp:libero}, our model and the publicly released
Stanford miniVLA checkpoint were trained on differently oriented
images. Our LIBERO\_90 dataset uses a 180-degree rotation
(\texttt{img[::-1, ::-1]}) of the raw simulator output, while the
canonical Stanford \texttt{modified\_libero\_rlds} convention uses
a vertical flip (\texttt{np.flipud}). To ensure fair head-to-head
evaluation, the inference harness applies a per-model orientation
transform at evaluation time: ROT180 for our model, FLIPUD for the
Stanford checkpoint.

We verified the required orientation per model with a small probe
on task 2 (5 trials per orientation, Stanford checkpoint). Success
rates under the four candidate transforms were: \texttt{none}:
$0/5$; \texttt{flipud}: $5/5$; \texttt{fliplr}: $0/5$;
\texttt{rot180}: $0/5$. This confirms that the Stanford checkpoint
requires FLIPUD at evaluation time, and is consistent with the
training-time orientation in
\texttt{modified\_libero\_rlds}.

\section{$\pi_{0.5}$ Fine-Tuning Details}
\label{app:impl:pi05}

\subsection{Base Model and Variants}
\label{app:impl:pi05:base}

We fine-tune $\pi_{0.5}$~\cite{black2025pi05}, a flow-matching
vision-language-action (VLA) model with approximately 2.3B
parameters. The model couples a SigLIP-So400m vision encoder with a
Gemma-2B backbone (initialized from PaliGemma) for joint language
and manipulation reasoning, together with a 311M-parameter
Gemma-based action expert that produces action chunks. We
initialize from the publicly released \texttt{pi05\_droid}
checkpoint, which was pre-trained on the full DROID corpus with
knowledge insulation. We selected this variant over the
LIBERO-pretrained checkpoint because the DROID pretraining yields
stronger transfer to our xArm7 real-robot setup: DROID was
collected across a diverse set of real-world manipulation scenes
and embodiments, whereas LIBERO is simulation-only, and we observed
in preliminary runs that the DROID initialization adapts more
readily to our overhead-plus-wrist RealSense camera configuration
and to the contact-rich dynamics of the real arm.

\subsection{LoRA Fine-Tuning}
\label{app:impl:pi05:lora}

We fine-tune with LoRA~\cite{hu2021loralowrankadaptationlarge}
rather than performing a full-parameter update, for three reasons.
First, each task in our suite has up to 100 teleoperated episodes
(roughly 13--28k frames), and full fine-tuning at this data scale
overfits before the policy meaningfully adapts; LoRA on a strong
DROID-pretrained prior preserves the language and visual backbones
and only adapts the manipulation-relevant subspace. Second, the
LoRA configuration fits comfortably on our compute budget, whereas
a full $\pi_{0.5}$ update requires sharded multi-GPU training with
FSDP and substantially longer wall-clock per step. Third, LoRA
keeps the base model byte-identical across the variants we compare,
which removes a confound when attributing performance differences
to the tactile overlay rather than to drift in the visual or
language backbone.

LoRA adapters are applied to both the 2B backbone and the 300M
action expert, on both the attention and feed-forward projections.
We use rank $r=16$ with $\alpha=16$ on the backbone, and rank
$r=32$ with $\alpha=32$ on the action expert, reflecting the larger
relative update we expect to be needed in the action-generation
pathway. All base weights are frozen; only the LoRA adapters and
the scale/bias parameters attached to them are updated. EMA is
disabled, as is standard for LoRA training.

\subsection{Observation and Action Spaces}
\label{app:impl:pi05:obs}

Each observation consists of three $224\times 224$ RGB camera
streams: an agent-view camera, a wrist-mounted camera, and a
zero-padded slot (with its image mask set to false) that matches
the dual-wrist convention of the base model since our xArm7 has a
single wrist camera. The proprioceptive state is an 8-dimensional
vector $[\,\text{ee\_pos} \in \mathbb{R}^3,\ \text{ee\_axis\_angle}
\in \mathbb{R}^3,\ \text{grasp},\ \text{grasp}\,]$, where the
grasp signal is duplicated to match the two-finger qpos convention
inherited from the base model's training data. The action is a
7-dimensional vector $[\,\Delta\text{xyz} \in \mathbb{R}^3,\
\Delta\text{axis\_angle} \in \mathbb{R}^3,\ \text{grasp} \in
\{-1,+1\}\,]$, expressed as world-frame deltas from the current
end-effector pose: $\text{xyz}_{t+1} = \text{xyz}_t +
\Delta\text{xyz}$ and $R_{t+1} = \Delta R \cdot R_t$
(left-multiplication). The policy predicts an action horizon of 10
steps per forward pass, corresponding to 1.0\,s of future control
at our 10\,Hz control rate.

\subsection{Hyperparameters}
\label{app:impl:pi05:hparams}

All 32 fine-tuning runs (4 tasks $\times$ 8 input
variants) share identical hyperparameters;
only the input variant differs.
Table~\ref{tab:pi05-hparams} summarizes the configuration.

\begin{table}[h]
\centering
\small
\caption{$\pi_{0.5}$ LoRA fine-tuning hyperparameters. All 32
runs share these settings; only the input variant differs.}
\label{tab:pi05-hparams}
\begin{tabular}{ll}
\toprule
Setting & Value \\
\midrule
Base checkpoint & \texttt{pi05\_droid} \\
Fine-tuning method & LoRA (backbone $r{=}16$, action expert $r{=}32$) \\
LoRA targets & attention and FFN projections \\
Image resolution & $224 \times 224 \times 3$, 3 cameras \\
Action horizon & 10 steps (1.0\,s at 10\,Hz) \\
Batch size & 8 \\
Optimizer & AdamW \\
Gradient clipping & global norm $\leq 1.0$ \\
LR schedule & cosine with linear warmup \\
\quad warmup steps & 500 \\
\quad peak LR & $1\times 10^{-4}$ \\
\quad decay steps & 20{,}000 \\
\quad final LR & $1\times 10^{-5}$ \\
EMA & disabled \\
Maximum training steps & 20{,}000 \\
Checkpoint interval & 2{,}000 steps \\
Early stopping & $<\!0.5\%$ relative improvement in the 1000-step \\
                & rolling-window mean loss for two consecutive \\
                & checks, after step 3{,}000 \\
\bottomrule
\end{tabular}
\end{table}

For context on the effective number of passes through the data: at
batch size 8, one epoch corresponds to roughly 1{,}600--3{,}500
optimizer steps depending on the task's episode lengths. Runs that
early-stopped in the 5{,}100--7{,}900 step range saw approximately
2--4 epochs; runs that ran to the full 20{,}000-step budget saw
approximately 6--12 epochs.

\subsection{Compute and Wall-Clock}
\label{app:impl:pi05:compute}

We trained on a SLURM-managed cluster of H200 GPUs, with each
fine-tuning run allocated 4$\times$H200 GPUs, 8 CPU cores, and
96\,GB of system memory, under a 24-hour walltime limit. The
32 task--variant combinations were trained in parallel across
the cluster. Each run early-stopped well within the walltime
budget, with most runs terminating between 5{,}100 and 7{,}900
steps and the longest-running configurations reaching the full
20{,}000-step cap when their loss curves continued to descend past
the early-stop checkpoint window.

\section{Real-World Experiment Details}
\label{app:realworld}

\subsection{Hardware Setup}
\label{app:realworld:setup}

Our real-world workspace consists of a UFactory xArm7 7-DoF
manipulator with a parallel-jaw gripper, on which we mount our
custom tactile sensors (one per finger; see
Appendix~\ref{app:hardware}). Visual observations come from two
Intel RealSense D435 cameras: one wrist-mounted and one
third-person external view. Cameras and tactile sensors are
spatially calibrated for overlay rendering via the forward
kinematics chain described in Appendix~\ref{app:projection}.

Raw magnetometer readings are normalized before overlay rendering
using a two-step procedure applied independently to each
magnetometer-axis channel. First, a constant resting-state offset is
subtracted: the sensor reading recorded at the start of a session with
no contact applied is taken as the per-channel baseline and subtracted
from all subsequent readings, removing the DC bias due to the sensor's
static magnetic environment. Second, each baseline-corrected channel is
scaled by the maximum absolute value observed for that same channel
across the demonstration dataset, mapping its dynamic range to
$[-1, 1]$. These per-channel scale factors are computed once from the
demonstration dataset and fixed thereafter. The per-channel baseline
offsets are re-recorded at the start of each episode or evaluation
rollout.

\subsection{Demonstration Collection}
\label{app:realworld:demos}

We collect 100 demonstrations per task using TeleDex~\cite{rayyan2026teledexaccessibledexterousteleoperation}, a smartphone-based teleoperation interface that streams the phone's 6-DoF pose (estimated on-device via ARKit) to the robot as relative end-effector commands. TeleDex requires no external tracking hardware or per-session calibration. During teleoperation we log raw streams from all sensors (RGB, proprioception, raw magnetic flux, gripper state) and apply our input variants \emph{post-hoc} from the same collected dataset, enabling controlled comparisons across input variants without re-collecting data.

\paragraph{Episode statistics.} Each demonstration is recorded at the robot's $10\,\mathrm{Hz}$ control rate. Per-task mean episode length and duration are summarized in \autoref{tab:demo-stats}. Mean episode duration ranges from \SI{13.1}{\second} for the single-stage Lift task to \SI{28.1}{\second} for Put Mug in Dishwasher. Across all 397 accepted demonstrations, the mean episode length is $210 \pm 74$ steps ($21.0 \pm 7.4\,\mathrm{s}$).


\paragraph{Acceptance criterion.} We accept a demonstration into the training set if (i) the operator performed at least one gripper open$\to$close transition during the episode (eliminating the occasional misfire where the operator forgot to grasp); and (ii) at least 5 non-paused control steps remain after we trim leading idle frames (defined as frames where the per-step translation $<0.5\,\mathrm{mm}$ \emph{and} per-step rotation $<0.5\,^{\circ}$). The criterion deliberately does not include a task-success label: TeleDex provides no in-the-moment success signal, and we found that incorporating an offline review pass biased the dataset toward shorter, less recoverable trajectories. With this filter, 397 of the 400 collected demonstrations ($99.25\%$) were used for training; the three rejected episodes were all from \textit{tube}, where the operator paused the robot for an extended setup window and never executed motion. The remaining $\sim$0.75\% loss has no measurable effect on downstream training-loss convergence (the smallest task, \textit{cube}, retains its full 100 demonstrations).

\begin{table}[h]
\centering
\caption{Per-task demonstration statistics (post-acceptance, post-idle-trim). $n$ is the number of episodes used in training; length is reported as mean $\pm$ standard deviation in control steps and seconds. Min/max bracket the full distribution.}
\label{tab:demo-stats}
\begin{tabular}{lcccc}
\toprule
Task        & $n$ & Length (steps)       & Duration (s)      & Min/Max (steps) \\
\midrule
cube        & 100 & $131 \pm 86$         & $13.1 \pm 8.6$    & 90 / 972 \\
tube        &  97 & $190 \pm 14$         & $19.0 \pm 1.4$    & 166 / 238 \\
charger     & 100 & $238 \pm 28$         & $23.8 \pm 2.7$    & 182 / 329 \\
dishwasher  & 100 & $281 \pm 32$         & $28.1 \pm 3.2$    & 236 / 554 \\
\midrule
\textbf{All} & \textbf{397} & $\mathbf{210 \pm 74}$ & $\mathbf{21.0 \pm 7.4}$ & 90 / 972 \\
\bottomrule
\end{tabular}
\end{table}

\subsection{Evaluation Protocol}
\label{app:realworld:eval}

Each condition is evaluated with 30 trials per task. Multi-stage
tasks are scored at the sub-task level; \cref{fig:realworld_main}
in the main text plots these results, and
Table~\ref{tab:realworld_main} reports the exact successes over 30
trials per sub-task.

\begin{table}[h]
\small
\centering
\caption{Real-world evaluation of $\pi_{0.5}$ fine-tuning across
four contact-rich manipulation tasks (30 trials per condition,
sub-task scored). Final-stage success corresponds to complete end-to-end task success. Binary Contact (1x) and (9x) denote aggregate per-finger and per-taxel binary contact, respectively.
}
\label{tab:realworld_main}
\renewcommand{\arraystretch}{1.2}
\begin{tabularx}{\linewidth}{ll Y YY YYY YY}
\toprule
& & \textbf{Lift} & \multicolumn{2}{c}{\textbf{Transfer Tube}} & \multicolumn{3}{c}{\textbf{Put Mug in Dishwasher}} & \multicolumn{2}{c}{\textbf{Plug Charger}} \\
\cmidrule(lr){3-3} \cmidrule(lr){4-5} \cmidrule(lr){6-8} \cmidrule(lr){9-10}
\multicolumn{2}{l}{\textbf{Method}} & Pick & Pick & Insert & Pull & Pick & Place & Pick & Insert \\
\midrule
\multicolumn{2}{l}{Baseline} & 20/30 & 5/30  & 0/30  & 21/30 & 11/30 & 11/30 & 9/30  & 0/30 \\
\multicolumn{2}{l}{Tac-View} & 16/30 & 9/30  & 0/30  & 22/30 & 18/30 & 16/30 & 17/30 & 0/30 \\
\multicolumn{2}{l}{Position-Only} & 22/30 & 21/30  & 7/30  & 17/30 & 14/30 & 11/30 & 9/30 & 0/30 \\
\midrule
\multirow{3}{*}{\ACRO}
 & \scriptsize{Binbars}      & 23/30          & 10/30          & 1/30           & 24/30          & 15/30          & 15/30          & 9/30           & 0/30          \\
 & \scriptsize{Binary Contact (1x)}      & 19/30          & 17/30          & 2/30           & 13/30          & 5/30          & 4/30          & 16/30           & 0/30          \\
 & \scriptsize{Avg.\ Arrow}  & 23/30          & 17/30          & 3/30           & 23/30          & 14/30          & 12/30          & 13/30          & 0/30          \\
  & \scriptsize{Binary Contact (9x)}      & 9/30          & 6/30          & 0/30           & 25/30          & 13/30          & 12/30          & 11/30           & 0/30          \\
 & \scriptsize{Multi-arrows} & \textbf{24/30} & \textbf{24/30} & \textbf{16/30} & \textbf{28/30} & \textbf{22/30} & \textbf{21/30} & \textbf{18/30} & \textbf{2/30} \\
\bottomrule
\end{tabularx}
\end{table}
 Reset between trials follows a fixed protocol in which the
object is placed at a pose sampled uniformly from a pre-marked
distribution region.

\paragraph{Step budget and action scale.}
Table~\ref{tab:eval_budget} summarizes the per-task rollout parameters.

\begin{table}[h]
\centering
\caption{Rollout budget and action scale per real-world task.}
\label{tab:eval_budget}
\setlength{\tabcolsep}{6pt}
\begin{tabular}{lcc}
\toprule
Task & Step budget & Action scale \\
\midrule
Lift              & 150 & 0.8 \\
Transfer Tube     & 400 & 0.4 \\
Put Mug in Dishwasher & 600 & 0.8 \\
Plug Charger      & 600 & 0.6 \\
\bottomrule
\end{tabular}
\end{table}

\paragraph{Per-stage success criteria.}
For single-stage tasks (Lift), a trial is scored successful if the object is lifted above a threshold height. For multi-stage tasks, each sub-stage is scored independently: in Transfer Tube, Pick is credited when the tube clears the holder and Insert is credited when it is placed at the target location; in Put Mug in Dishwasher, Pull is credited when the rack is pulled out to the target region, Pick is credited when the mug is lifted and held stably, and Place is credited when the mug is seated in the rack; in Plug Charger, Pick is credited when the charger is held securely and Insert is credited when the plug is fully seated in the socket. A trial is terminated early and marked as failed if the robot collides with the object or table at any point.

\paragraph{Retry behavior.}
Within the step budget the policy is free to make multiple grasp attempts if earlier attempts miss. For all tasks except Transfer Tube, the policy re-attempts after a missed grasp regardless of whether tactile observations are present. The Transfer Tube Pick stage is an exception: with tactile observations the policy re-attempts after a miss, but without tactile observations it does not, and the trial expires unused. We attribute this to the perceptual difficulty of the transparent tube: the tube's visual appearance provides less reliable grasp-failure signal, so without tactile feedback the policy cannot distinguish a missed grasp from other low-contact states and does not enter a recovery regime. With tactile feedback, the signal after a missed grasp unambiguously identifies the failure and prompts re-attempt. This emergent retry behavior is a qualitative benefit of contact sensing that goes beyond raw success-rate gains.

\end{document}